%% file: main.tex
\documentclass[fleqn,10pt]{wlscirep}
\usepackage[utf8]{inputenc}
\usepackage[T1]{fontenc}

\usepackage{subcaption}

\usepackage{float}

\usepackage{placeins}

\usepackage{comment}

\usepackage{booktabs}
\usepackage{multirow}
\usepackage{listings}
\makeatletter
\newcommand{\fakesub}[1]{%
  \par\vspace{6pt}%
  {\noindent\sffamily\bfseries #1}\par\vspace{0pt}%
  \@afterindentfalse\@afterheading%
}
\makeatother

\makeatletter
\newcommand{\fakesec}[1]{%
  \par\vspace{8pt}%
  {\noindent\sffamily\bfseries\large #1}\par\vspace{4pt}%
  \@afterindentfalse\@afterheading%
}
\makeatother

\title{
Large language models replicate and predict human cooperation across experiments in game theory
}

\author[1]{Andrea Cera Palatsi}
\author[2]{Samuel Martin-Gutierrez}
\author[3]{Ana S. Cardenal}
\author[1,4*]{Max Pellert}
\affil[1]{Department for Computational Social Sciences and Humanities, Barcelona Supercomputing Center}
\affil[2]{Grupo de Sistemas Complejos, Universidad Politécnica de Madrid}
\affil[3]{School of Law and Political Science, Universitat Oberta de Catalunya}
\affil[4]{Complexity Science Hub}

\affil[*]{max.pellert@bsc.es}

\begin{abstract}
Large language models (LLMs) are increasingly deployed as decision-making agents in high-stakes domains and as imitators of human behavior in the social and behavioral sciences. Yet how closely LLMs mirror actual human decision-making remains poorly understood. This gap is critical: misalignment could produce harmful outcomes in practical applications, while failure to replicate human behavior renders LLMs ineffective as social simulators. Here, we address this gap by replicating large-scale game-theoretic experiments and by introducing a systematic prompting and probing framework for machine-behavioral evaluation. We test three open models that are typically used to power agents (Llama, Mistral, and Qwen). We find that, across 121 dyadic games spanning four classical game types, Llama reproduces human cooperation patterns with high fidelity, while Qwen aligns closely with Nash equilibrium predictions. Characterizing models through behavioral phenotyping, we find that humans and Llama share an envious decision profile, while Qwen is more balanced overall and Mistral most strongly displays the trustful and optimist archetypes. An attention-based analysis of payoff salience reveals that Llama processes payoff information in a structured, layer-dependent manner that is absent in Qwen and Mistral, suggesting a mechanistic basis for its closer alignment with human behavior. Notably, population-level behavioral replication is achieved without persona-based prompting, simplifying the simulation process. Extending the experimental parameter space beyond the original human-tested games, we generate and preregister testable hypotheses for novel game configurations. Our findings demonstrate that appropriately configured LLMs can replicate aggregate human behavioral patterns, exhibit human-like decision phenotypes, and enable systematic exploration of unexplored experimental spaces, offering a complementary approach to traditional behavioral research that generates new empirical predictions about human social decision-making.
\end{abstract}
\begin{document}

\flushbottom
\maketitle

\thispagestyle{empty}

The capabilities of large language models (LLMs) have made their use as human behavioral imitators a subject of active study and a tool for inference in the social and behavioral sciences. LLMs as computational complements to humans offer the promise of first replicating experiments originally conducted with humans and then extending them to previously unexplored areas, making the experimental process much faster and cheaper while operating under less stringent ethical constraints. The algorithmic fidelity\cite{argyleOutOneMany2023a} of these synthetic approaches hinges on their ability to robustly elicit valid behaviors from the models. While training data potentially contains vast implicit information on human patterns of experience and behavior, researchers face many modeling choices and challenges in their attempts to reliably extract those patterns. To simulate social experiments, we first need a systematic evaluation framework to assess whether LLMs can replicate empirically observed human patterns.

Here, we contribute such a framework by replicating large-scale, generalized experiments in game theory with open LLMs, comparing their choices against both empirical human behavior and formal game-theoretic benchmarks. We use this setting to ask three related questions: whether LLMs reproduce aggregate human cooperation patterns across a diverse set of games, whether they exhibit stable behavioral phenotypes analogous to those observed in humans, and whether their internal processing of payoff information can help explain behavioral differences.

For humans, game-theoretic experiments have established the presence of distinct behavioral phenotypes of decision-making across a generalized framework covering several classical game scenarios \cite{poncela2016humans}. By comparing the behavior of LLMs under the same conditions as those originally used with human participants, we gain valuable insights into machine behavior \cite{rahwanMachineBehaviour2019b}. When LLM-powered agents\cite{wangSurveyLargeLanguage2024} make decisions, it is crucial to understand whether there are systematic behavioral differences across model types that may influence their choices. Should LLMs display behavioral phenotypes analogous to those observed in humans, the implications for agentic AI making decisions across high-stakes domains\cite{bommasaniOpportunitiesRisksFoundation2021a} such as medicine\cite{thirunavukarasu_2023, singhalLargeLanguageModels2023}, law\cite{katzGPT4PassesBar2024}, and education\cite{chirikovGenerativeAIUse2026} could be far-reaching. Certain LLMs could lead agents to imitate human social decision-making, while others may exhibit systematic deviations from human norms.

Beyond comparing choices, we also seek to understand why some models may align more closely with human behavior than others. To probe those differences more deeply, we need to further our understanding of the internal information-processing patterns of LLMs. Distinct representational features uncovered in these models could in turn inform our understanding of human social decision-making, where we are faced with far more fundamental ethical and biological limits to inspecting the underlying mechanisms. The realism of such attempts to use LLMs as "cognitive models" \cite{binzFoundationModelPredict2025} remains debated \cite{xieCentaurMayHave2025} and part of observed model performance may be due to Clever Hans effects \cite{sturmSimpleMethodDetermine2014}. Given these concerns, rather than attempting to simulate the micro-level of individual participants, a goal that presents substantial challenges, we focus on the more tractable objective of capturing macro-level behavioral patterns across participants and types of experiments. This approach, while not without its own challenges, provides an effective means for understanding collective human behavior. Replicating existing human experimental setups is only the first step: ultimately, we want to derive empirically testable hypotheses about novel experimental configurations that can guide future human experiments and inform theory-building.

To pursue these goals, we need modeling frameworks that are both tractable and tightly constrained so as to manage the wilderness that LLM-based modeling presents. The substantial freedom that comes with LLM-based modeling makes it essential to work within well-defined domains where approaches can be evaluated clearly. The scenarios that (algorithmic) game theory \cite{roughgardenAlgorithmicGameTheory2010} provides are well-suited for this purpose, as they offer 1) analytical (often "optimally rational") solutions, 2) ample empirical data of humans playing such games and 3) mathematically well-defined, simple tasks with clear instructions that can be readily translated to natural language. Furthermore, despite their simplicity, these game-theoretic scenarios are ubiquitous and offer powerful analytic tools for the kinds of complex social phenomena relevant to contexts where agentic AI is increasingly deployed. 

\subsection*{Examples of LLMs playing games}

Existing research has examined LLMs in various game theoretical settings including the Dictator, Ultimatum, Trust Game and the Prisoner's dilemma as well as classic behavioral experiments such as Wisdom of Crowds and the Milgram obedience experiment (for details see \ref{app:subsec:detailgames}).

The earliest work in this domain tested LLMs (different models from the GPT series) with the Dictator Game as well as with other decision-making scenarios \cite{horton_2023}. When given no persona instructions, the most advanced model tested (GPT-3 text-davinci-003) consistently selected efficiency-maximizing outcomes and did not vary its responses to the same game when asked repeatedly. The authors instructed LLMs in different conditions to be inequity-averse (care about fairness between players), efficient (care about the total payoff of both players) and selfish (care about their own payoff). The models responded by following these instructions precisely when capable, though less advanced GPT-3 models could not modulate their behavior based on instructions. When the model (GPT-3 text-davinci-003) received no instructions about how to value fairness, efficiency, or self-interest, it consistently chose efficient outcomes that maximized total payoff. However, to replicate actual human behavior patterns, a synthetic population needed to be composed of 52\% selfish agents, 32\% efficiency-focused agents, and only 15\% inequity-averse agents, suggesting that the model's default preferences, inclined towards efficiency and increasing common benefit, differ from human tendencies, mostly associated with selfish behaviors. 

The same authors also simulated a scenario in which a store increased the price of snow shovels after a snowstorm and found that LLMs were more likely to reject larger price increases. Prompting models to have different political ideologies changed their responses (e.g. left-wing personas consistently rated price increases as unfair, while moderate and libertarian personas found smaller increases acceptable). Another scenario had models choosing between different options with one that was presented as the status quo. Models, similar to humans, preferred the status quo. A final scenario had models acting as employers choosing between candidates with different experience levels. When a minimum wage was imposed, LLMs shifted towards hiring more experienced applicants, a pattern also found among humans.

A similar line of research tested eight GPT model variants, from text-ada-001 to GPT-4 \cite{aher_2024_simulating}. In the Ultimatum Game, text-davinci-002 did not accept unfair proposals below 10\% of the endowment, and showed high acceptance rates for offers of 50\% or more, a gradual pattern that appears in humans, showing behavior consistent with human tendencies to reject unfair proposals. LLMs also played the Milgram obedience experiment, with researchers creating an alternative "sleepiness and driving" scenario to prevent the models from replicating memorized descriptions of the original experiment. In the end, text-davinci-002 displayed the same tendency of obedience that was originally observed in humans, though at slightly higher rates (75\% vs. 65\% in the original human study). While text-davinci-002 successfully replicated human behavioral patterns in these experiments, more recent models like GPT-4 were only tested in the Wisdom of Crowds study, where they exhibited "hyper-accuracy distortion" by providing inhumanly accurate answers (often perfectly correct) rather than the varied, imperfect estimates typical of humans. This prevented the Wisdom of Crowds aggregation effect from functioning properly, revealing how alignment procedures that improve factual accuracy can create systematic differences from human cognition.

Mei et al. \cite{mei_2024} had LLMs play games and administered the Big 5 personality questionnaire to the models. Apart from measuring behavioral traits, they performed Turing experiments, selecting individual responses to games and comparing them with human responses. For the Big 5 personality test, they compared the answers of two models with human distributions. They confirmed that responses of the same model did not present much variation, similar to the consistency a single human subject would show if repeatedly queried. The Big 5 results of the advanced models were the most similar to humans, to the point of becoming statistically indistinguishable. For the games, GPT-4 passed the behavioral Turing test in most games, but showed detectably different patterns in the Trust Game, where it demonstrated greater trust and generosity than typical humans, consistently avoiding purely selfish strategies (like giving \$0) that humans frequently chose. Humans exhibited more extreme and varied actions across the behavioral spectrum. In the Prisoner's Dilemma, it was more cooperative than the median human, who showed greater inclination to defect. The advanced model showed adaptive behavior, modifying its responses based on previous experience "as if" learning from interactions. In the Prisoner's Dilemma, GPT-4 modified its responses depending on the past, showing conditional cooperation strategies similar to those observed in humans.

In summary, existing research on LLMs playing classical game-theoretic scenarios yields mixed evidence, with a few recurring patterns. Advanced models can exhibit behavioral patterns similar to humans in experimental settings, though when differences arise, LLMs typically demonstrate greater cooperation than human participants. For some experimental setups, a key methodological challenge is that each model instance effectively acts as a single subject, requiring deliberate strategies to link the diversity observed in human populations on the micro-level to macro-level outcomes\cite{asheryEmergentSocialConventions2025}. While prompts building on extended interviews with humans show the greatest potential for endowing LLMs with distinct personalities, simpler approaches using only demographic attributes or brief descriptions can also be used to generate variation. Notably, in some studies, LLMs tend to attribute higher knowledge and accuracy to others than is empirically typical (a form of hyper-accuracy bias), and advanced models often exhibit payoff-maximizing strategies that benefit both players. Finally, most existing studies rely on commercial models accessed via closed, black-box APIs, limiting transparency on model changes, as well as feasibility of analysis of model internals and reproducibility.

\subsection*{Choice of models}

We focus on widely used, open-source models that are both archetypal of current LLM capabilities and likely to power agents. We select three medium-sized models that balance performance with computational feasibility for typical agentic AI pipelines while ensuring architectural diversity. We derive our main results using \textbf{Llama-3.1-8B-Instruct} by Meta AI\cite{grattafioriLlama3Herd2024}, as justified in Section \ref{sec:results}. To assess robustness across model architectures and uncover patterns of behavior across models, we repeat the exact same analyses with \textbf{Mistral-7B-Instruct-v0.3} by Mistral AI\cite{jiangMistral7B2023} and \textbf{Qwen2.5-7B-Instruct} by Alibaba Cloud\cite{yangQwen2TechnicalReport2024,qwenQwen25TechnicalReport2025} in the Appendix. Additionally, we employ Qwen2.5-7B-Instruct for the separate tasks of answer extraction and logical verification (for details see Methods Section \ref{subsec:method_extract}). We refer to these models as Llama, Mistral, and Qwen.

\subsection*{Simulating game theoretical experiments}

We base our LLM setup on the experiments of Poncela-Casasnovas et al. \cite{poncela2016humans}, who collected data from more than five hundred participants playing dyadic games. Participants were recruited from the general audience of the game festival DAU Barcelona\cite{vicensCitizenSocialLab2018} to play games with random opponents and payoff configurations randomly selected from a pre-set parameter grid. Experimenters paid participants with lottery tickets according to the points they scored. Each game consisted of a single round in which both players made one decision simultaneously, without receiving feedback about the opponent's choice until after both had decided.

Each player could choose to cooperate or defect (encoded as colors to prevent wording bias for the human participants). Depending on the choices of both players, there are four possible outcomes (Table \ref{table_payoff}).
If a player chooses C, they receive payoff $R$ if the opponent also chooses C, or payoff $S$ if the opponent chooses D. If a player chooses D, they receive payoff $T$ if the opponent chooses C, or payoff $P$ if the opponent also chooses D.

\renewcommand{\arraystretch}{1.3}

\begin{table}[htbp!]
\caption{\textbf{Payoff matrix for strategies C (Cooperate) and D (Defect).} Each cell shows the payoff as $(\text{Player 1}, \text{Player 2})$. The parameters represent: $R$: \textbf{r}eward for mutual cooperation; $P$: \textbf{p}unishment for mutual defection; $T$: \textbf{t}emptation to defect; $S$: \textbf{s}ucker’s payoff.}
\label{table_payoff}
\centering
\begin{tabular}{cccc}
 & & \multicolumn{2}{c}{\textbf{Player 2}} \\
 & & C & D \\
\cline{3-4}
\multirow{2}{*}{\textbf{Player 1}} 
 & C & \multicolumn{1}{|c|}{$(R,R)$} & \multicolumn{1}{c|}{$(S,T)$} \\
\cline{3-4}
 & D & \multicolumn{1}{|c|}{$(T,S)$} & \multicolumn{1}{c|}{$(P,P)$} \\
\cline{3-4}
\end{tabular}
\end{table}

In the experiments, the value of $R$ was kept constant at 10 points and the value of $P$ at 5 points, ensuring that mutual cooperation always yields a higher payoff than mutual defection. The parameter $S$ ranged from 0 to 10, while $T$ ranged from 5 to 15, with both taking integer values. This created 11 possible values for each parameter, leading to 121 distinct games.

These parameter combinations constitute a generalized framework that encompasses four classical game types:

\begin{itemize}
   \item HG (Harmony Game): $S > P$, $R > T$, here $S > 5$ and $T < 10$. Cooperation is the dominant strategy.
   \item SG (Snowdrift Game): $T > R > S > P$, here $T > 10 > S > 5$. The best individual outcome is to defect when the opponent cooperates, but the worst outcome is mutual defection.
    \item SH (Stag Hunt Game): $R > T \ge P > S$, here $10 > T \ge 5 > S$. The best outcome is mutual cooperation, but the worst outcome is to cooperate when the opponent defects.
    \item PD (Prisoner’s Dilemma): $T > R > P > S$, here $T > 10 > 5 > S$. The best individual outcome is to defect when the opponent cooperates, and the worst is to cooperate when the opponent defects.
 \end{itemize}

To analyze human participants' behavior across games with uniformly randomly selected parameters, the authors used K-means clustering to identify 5 distinct behavioral phenotypes that characterize player strategies. Each of those phenotypes is characterized by specific, idealized decision rules:

\begin{itemize}
    \item \textbf{Optimist.} Maximizes the best possible outcome. Cooperates if $R > T$, assuming the opponent will cooperate.
    \item \textbf{Pessimist.} Maximizes the worst possible outcome (minimax strategy). Cooperates if $S > P$.
    \item \textbf{Envious.} Seeks to maximize relative advantage over the opponent. Cooperates if $S \ge T$.
    \item \textbf{Trustful.} Always cooperates, representing the strongest inclination toward maximizing joint payoffs.    
    \item \textbf{Undefined.} Exhibits no consistent strategy, cooperating and defecting randomly. 
\end{itemize}

We adopt and adapt this experimental framework for LLMs to identify cross-game cooperation patterns, characterize behavioral differences across models, and analyze their information-processing strategies. In addition to comparing LLM responses to empirical human cooperation rates, we also compare them to the game-theoretic optima given by the Nash equilibrium solution of each game (see Methods Section \ref{subsec:analytic}).

Instead of fully replicating the original experimental setup, we simplify it to focus on obtaining the main statistic we aim to reproduce: cooperation rates. We present each payoff configuration independently to each model and elicit a cooperation or defection choice, without simulating repeated play, pairing models against one another, or tracking accumulated payoffs. Full implementation details are provided in Methods Section \ref{subsec:method_adapt}.

Finally, we go beyond replication by introducing novel games designed to explore experimental configurations outside of the original experiments. In doing so, we employ LLMs as predictive proxies for human behavior, demonstrating their potential to forecast how humans might behave in unexplored strategic settings, which represents a key application of AI for scientific discovery\cite{evansScience2025}.

Our work builds upon and extends earlier attempts to subject LLMs to classical game-theoretic scenarios. While earlier studies are valuable as proof-of-concept demonstrations\cite{kitadaiCanAIHigh2025a}, they typically lack thorough calibration against empirical human data collected across different types of games. We advance this research program in several ways. First, we systematically validate our approach using multiple open-source models rather than relying on API access to a limited set of commercial models, enabling assessment of robustness across different architectures and the identification of cross-model behavioral patterns. Second, we develop and validate a rigorous prompting methodology informed by empirical testing, incorporating logical verifiers and other quality control mechanisms to ensure response validity. Third, we establish behavioral phenotypes of the models. Fourth, we leverage that we can fully access the open models we use locally to conduct analyses of internal model representations and establish payoff saliency profiles through attention analysis. Fifth, we use our workflows not merely to replicate known patterns, but to systematically expand the experimental parameter space, generating empirically testable hypotheses for novel games that can guide future human experiments and theory-building.

\section{\label{sec:results}Results}

For each model and each parameter combination, we repeatedly prompt the model to play the corresponding dyadic game and extract a binary cooperation or defection decision from each response. In the cooperation matrices that we show, each cell represents the average cooperation rate across repeated model decisions in a specific game defined by a particular combination of $S$ and $T$ values. As in the original experiments, we kept $R$ and $P$ constant at 10 and 5.

\subsection{Replication of human behavioral patterns through answer extraction refinement}

\begin{figure}
    \centering
    \includegraphics[width=\linewidth]{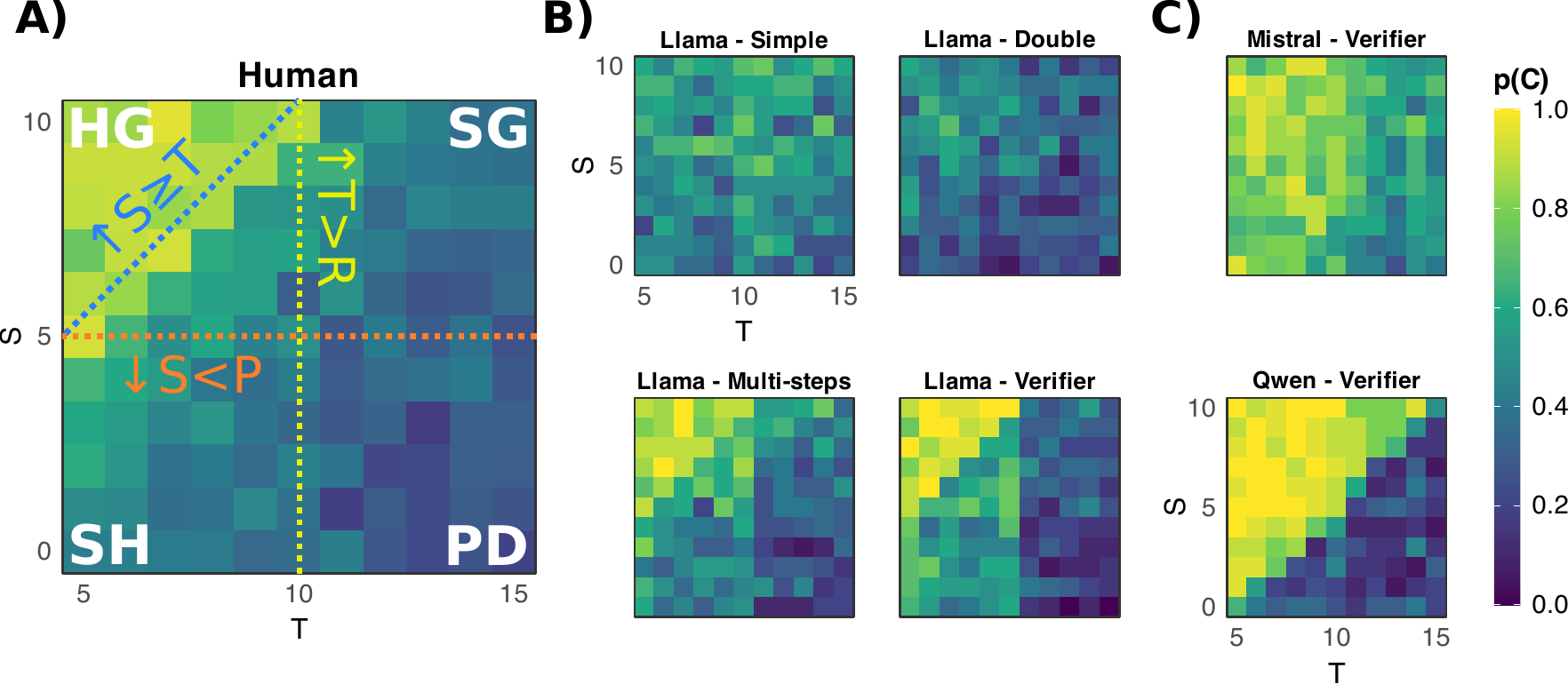}
    \caption{\textbf{Cooperation patterns of humans and LLMs.} Panel A: Human cooperation matrix from the original experiments\cite{poncela2016humans} with an average cooperation of 0.487. Each quadrant of the parameter grid corresponds to a standard game in game theory: HG (Harmony Game), SG (Snowdrift Game), SH (Stag Hunt Game), and PD (Prisoner’s Dilemma). Games outside these quadrants represent non-standard game configurations. Panel B: Progressive improvements in revealing human empirical patterns in Llama's cooperation matrix as extraction complexity increases. (For the full set of plots of the other two models, see Figure \ref{app:fig:methodsmistralqwen}.) At the final stage with the Logical Verifier, Llama shows an average cooperation of 0.492. Panel C: Average cooperation at the final stage is 0.682 for Mistral and 0.577 for Qwen (Nash would predict 0.500). Color scales indicate average cooperation from 0 (purple: no cooperation) to 1 (yellow: full cooperation).}
    \label{fig:refinement_models}
\end{figure}

We develop four progressively more complex methods to elicit cooperation or defection decisions from the models (for details see Methods Section \ref{subsec:method_extract}):

\begin{itemize}
    \item \textbf{Simple extraction.} After providing the instructions, we directly ask the LLMs to state their decision by generating only the corresponding label ($A$ for defection and $B$ for cooperation, or the other way round). 
    \item \textbf{Double extraction.} We ask the LLMs for longer, free-text answers, then use Qwen to extract the decision from this longer response. 
    \item \textbf{Multi-step extraction.} We ask the LLMs for a long answer using a prompt that guides models through logical reasoning steps, then use Qwen to extract the decision. 
    \item \textbf{Logical verifier extraction.} We ask the LLMs for a long answer using a prompt that guides the model through logical reasoning steps. Before extraction, we apply Qwen as a logical verifier to filter the responses (for details see Subsubsection~\ref{subsec:logicalverifier}), and then use Qwen again to extract the decision. 
\end{itemize}

Figure \ref{fig:refinement_models} demonstrates visually how each step of our refinement procedure progressively reduces noise in the cooperation patterns of the Llama model, revealing a pattern that closely resembles the empirical human cooperation matrices (Panel A): First, when using only "Simple Extraction", cooperation appears almost random. Second, upon using "Double Extraction", we start observing the game regions forming, with decreased cooperation in the Prisoner's Dilemma region. Third, with "Multi-step Extraction", we see two distinct regions emerge: games where $S \ge T$ exhibit increased cooperation (with highest cooperation in the Harmony Game ($S > 5$, $T < 10$)), while games where $T > R$ exhibit decreased cooperation. With the final layer of complexity, the logical verifier, Llama closely replicates the original empirical human patterns. At the "Logical Verifier" stage for the other two models (Panel C), Mistral displays overall high levels of cooperation with a vertical line that roughly separates the regions where $T > R$ (less cooperative) from regions where $T \leq R$ (more cooperative). Qwen closely follows Nash equilibrium (non-transparent inset in Panel B of Figure \ref{fig:novel_games}). Figure \ref{app:fig:methodsmistralqwen} shows all four stages for Qwen and Mistral. 

Llama (and humans) display distinct game-specific behavior: In the Harmony Game (HG) region there is a clear area defined by $S \ge T$ that displays distinctly high cooperation. In comparison, the quadrant below (Staghunt Game (SH) area with $T \leq R$) is defined by lower but still elevated cooperation. On the right of it, the Prisoner's Dilemma (PD) region shows least cooperation. The Snowdrift Game (SG) area is similarly characterized by decreased cooperation, but still slightly higher than PD. If we compare that to the Nash equilibrium cooperation matrix (non-transparent inset of Panel B in Figure \ref{fig:novel_games}), we find: The rational choice for HG is full cooperation. For SG, we find a mixed equilibrium, where both strategies are expected in different proportions. For SH, there is a boundary that separates two sub-regions depending on the stable strategy, while for PD the rational choice is to defect. For discussions of Mistral and Qwen, see Subsection \ref{app:subsec:othermodels} in the Appendix.

Figure \ref{fig:prompting_stages} quantifies these changes by showing Pearson's correlation and mean squared deviation with the human and Nash cooperation matrices for each model and each prompting stage. Multi-step instructions add most signal for Llama, while Qwen is mostly robust from double extraction onward. Mistral benefits monotonically from all four stages.
Our findings are robust to setting temperature above 0: as expected, increasing temperature from 0.1 to 1 generally leads to increasing variance of model cooperation rates (Figure \ref{fig:tempvariance}), while point estimates stay stable (Figure \ref{fig:tempperformance}).

\begin{figure}
    \centering
    \includegraphics[width=0.5\linewidth]{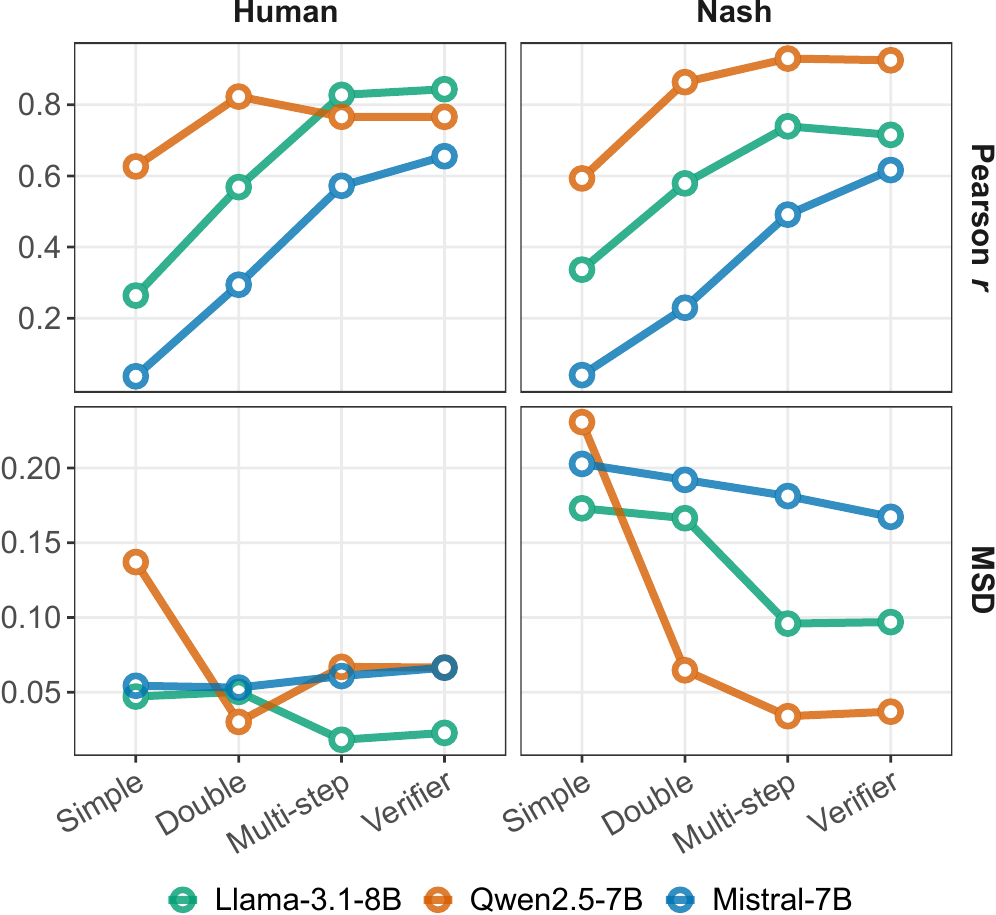}
    \caption{\textbf{Behavioral imitation across prompting stages.} Pearson's $r$ ($\uparrow$ = closer) and mean squared displacement (MSD) ($\downarrow$ = closer) with the human and the Nash cooperation matrix across all 121 games are shown for each model. Simple extraction produces the least signal for all models. Double extraction helps all models, with Mistral and Llama also substantially benefiting from multi-step extraction and the logical verifier.}
    \label{fig:prompting_stages}
\end{figure}

\subsection{Cross-model comparison}

\begin{figure}

\centering

  \begin{minipage}{0.44\textwidth}
    \centering
    \includegraphics[width=\linewidth]{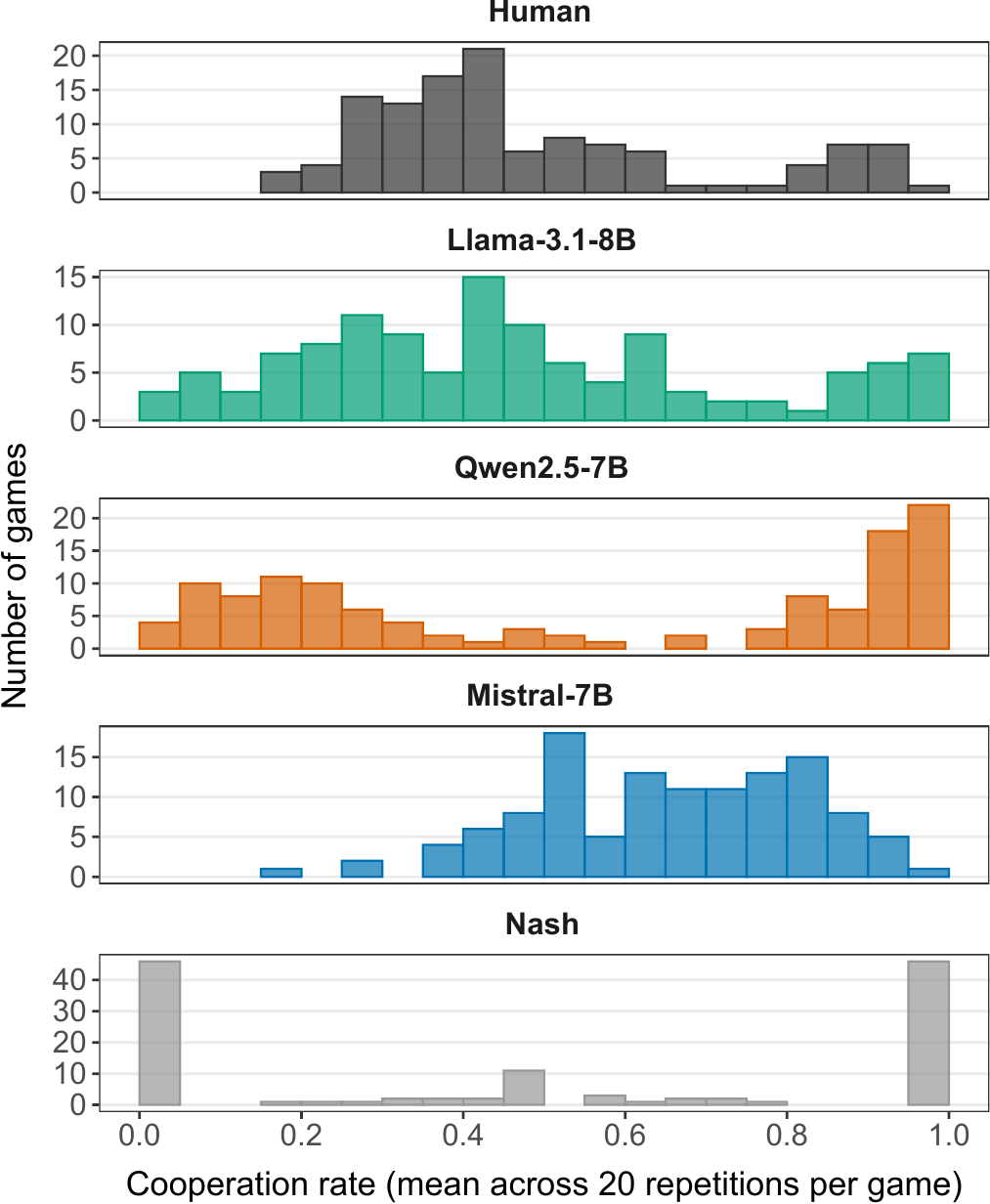}
    \caption{\textbf{Distribution of cooperation rates.} Llama is closest to the distribution of humans, while Qwen approximates the bimodal distribution of Nash. Mistral is least similar to any of the other distributions.}
    \label{fig:coopdist}
  \end{minipage}
  \hfill
  \begin{minipage}{0.53\textwidth}
    \centering
\input{metrics_table.tex}
    \captionof{table}{\textbf{Metrics for model comparisons with human data and Nash.} Mean squared displacement ($\downarrow$ = closer) and Pearson's $\boldsymbol{r}$ ($\uparrow$ = closer) comparing the cooperation matrices from all 121 games of the three LLMs with those of human participants and the Nash equilibrium predictions. Llama is closest to the human data, while Qwen exhibits the highest relationship to Nash. Compared to Llama, Nash and Qwen are less strongly associated with the empirical human behavioral patterns. Mistral displays an intermediate pattern, without a clear affinity for either of the two benchmarks. Pearson $r$ CIs use the Fisher z-transformation; MSD CIs use a normal approximation on the squared errors.}
    \label{combined_table}
  \end{minipage}
\end{figure}

Table \ref{combined_table} confirms that Llama has the highest similarity to humans (MSD = $0.020$, $r = 0.84$), showing non-overlapping confidence intervals with the second-most similar model, Qwen. Qwen exhibits the highest similarity to Nash equilibrium (MSD = $0.033$, $\boldsymbol{r} = 0.93$). Mistral is similar neither to humans nor the Nash equilibrium. Importantly, Llama replicates human cooperation patterns closer than the approach building on Nash equilibrium does (MSD = $0.095$, $\boldsymbol{r}=0.78$), suggesting that LLM-based behavioral modeling comes closer than classical rational choice predictions to capturing actual empirical human decision-making.

Comparisons of average cooperation rates over all 121 games show that humans  cooperate close to but slighly less (0.487) than Llama (0.492) that, in turn, cooperates slightly less than expected from Nash (0.500). Qwen cooperates a bit more (0.577) and Mistral substantially more (0.682). While average cooperation rates, with the exception of Mistral, are close to each other, the distributions of cooperation rates (Figure \ref{fig:coopdist}) show distinct features: Llama is clearly closest to the human distribution ($W_1 = 0.055$). Qwen is approaching the bimodality of Nash ($W_1 = 0.105$). Mistral has less similarity to any of the other distributions ($0.195 \leq W_1 \leq 0.297$).

\renewcommand{\arraystretch}{1.4}

\subsection{Behavioral phenotypes of models}

\begin{figure}
    \centering
    \includegraphics[width=\linewidth]{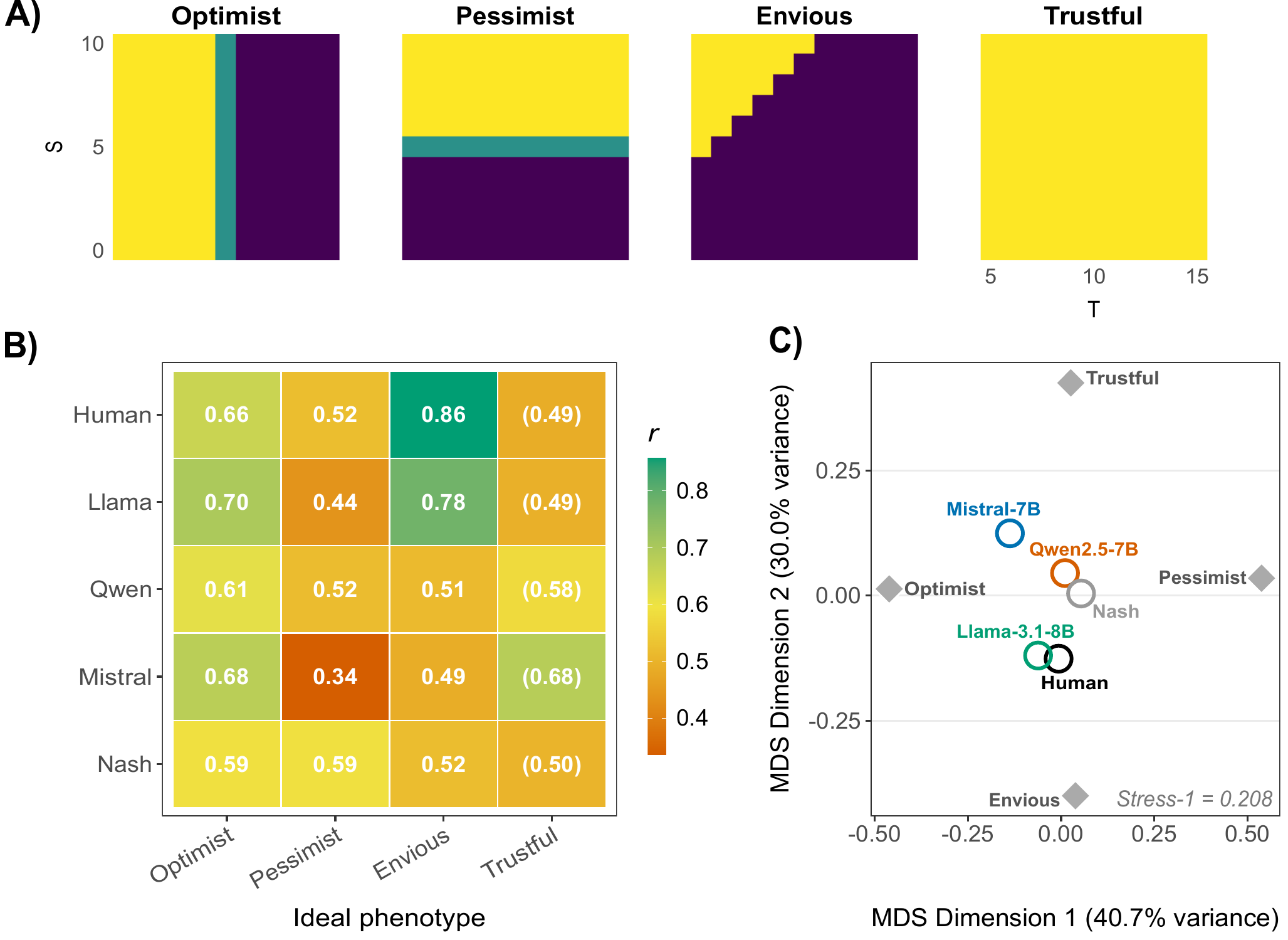}
    \caption{\textbf{Idealized phenotype cooperation matrices, heatmap of model, human, Nash vs. idealized phenotype cooperation and phenotype similarity in cooperation space} Panel A: Idealized cooperation matrices for the behavioral phenotypes identified in the original experiments with human participants\cite{poncela2016humans}. Panel B: Pearson correlation across all 121 games. Green = stronger pattern match. Red = weaker pattern match. (Trustful is constant; value shown is mean cooperation rate as the most suitable proxy.) Human and Llama are both most Envious. Qwen is most Optimist, but similar to Nash it is rather flat across phenotypes, confirming both not to resemble any single archetype. Mistral is highest Trustful and Optimist and lowest Pessimist. Panel C: Multidimensional scaling (MDS) of pairwise behavioral distances (computed from the data shown in Panel B) between cooperation patterns of models, humans and Nash (rings) and cooperation by ideal phenotypes (diamonds). Proximity indicates greater similarity in cooperation behavior across the game matrix. The near-overlap of Human and Llama reflects the highest empirical correlation among all model–human pairs.}
    \label{fig:ideal_phenotype_heatmap_mds}
\end{figure}

To characterize behavioral phenotypes of the models, we build the idealized cooperation matrix for each phenotype identified in the original experiment (Panel A of Figure \ref{fig:ideal_phenotype_heatmap_mds}) and quantify each model's proximity to these phenotypes. Panel B of Figure \ref{fig:ideal_phenotype_heatmap_mds} shows the correlation between each phenotype matrix and each model's cooperation matrix. The exception is the Trustful phenotype, for which we use average cooperation instead of the Pearson correlation used for the other phenotypes (see Methods Section \ref{subsec:phenotyping_methods}).

Humans are most strongly aligned with the Envious phenotype ($r = 0.86$), which maximizes relative advantage by cooperating only when $S \geq T$. Llama closely mirrors this profile ($r = 0.78$), consistent with its overall similarity to human cooperation patterns. Qwen shows the highest correlation with the Optimist phenotype ($r = 0.61$), but its correlations are relatively flat across phenotypes, as is Nash, confirming that neither resembles any single archetype strongly. Mistral shows the highest mean cooperation rate among models, consistent with a Trustful and Optimist tendency, and the weakest Pessimist similarity. Panel C of Figure \ref{fig:ideal_phenotype_heatmap_mds} shows the pairwise behavioral distances computed from the data in Panel B reduced to 2D through multidimensional scaling, in which Human and Llama nearly overlap, both positioned closest to the Envious phenotype. Consistent with Panel B, Qwen is placed closest to Nash, but a bit closer to Optimist and Trustful and a bit further form Pessimist, whereas Mistral is the closest model to Optimist and Trustful and the furthest from Pessimist.

\subsection{Payoff saliency differences between models}

\begin{figure}
    \centering
    \includegraphics[width=0.5\linewidth]{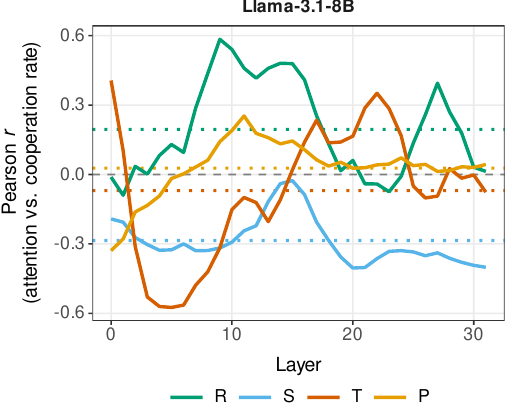}
    \caption{\textbf{Payoff salience: attention to payoff values vs. cooperation rate.} We extract attention of the last token of the instruction prompt that includes a payoff table to each of the four payoff values. We average over both prompt orderings per game, then correlate the resulting per-game attention values with cooperation rate across the 121 games. We use a rolling mean (window = 3) to display the lines (dashed lines show the mean). Llama exhibits a structured, layer-dependent payoff saliency profile in which R and T are anti-correlated across network depth, S is persistently negative, and P carries no consistent signal; neither Qwen nor Mistral replicate this pattern (Figure \ref{fig:attention_othermodels}).}
    \label{fig:attention_analysis}
\end{figure}

To test whether models that better reproduce human cooperation patterns also process payoff information differently, we compute the correlation between layer-wise attention to the tokens encoding each payoff value (R, S, T, P) and the cooperation rates observed across games. In Figure \ref{fig:attention_analysis} we present the results for Llama, which exhibits a clear structure across network depth. R shows a sustained positive correlation across most layers, peaking at $r \approx 0.6$ around layer 9-10. S is persistently negatively correlated with cooperation throughout all layers (mean $r \approx -0.3$). T shows the most striking pattern: mildly positive at layer 0, the correlation drops sharply to $r \approx -0.6$ by layers 4–5, then recovers toward zero and remains near zero through the later layers. Notably, the R and T profiles tend to move in opposing directions across layers: as T-attention drops sharply in the early layers, R-attention rises, with the two signals tracking each other inversely through approximately layer 15 before T recovers toward zero. P shows no consistent signal across layers (mean $r \approx 0.03$).

The opposing trajectories of R and T are theoretically coherent and suggest that Llama organizes payoff information around the central trade-off between mutual cooperation and unilateral defection. The sustained positive R signal peaking at mid-network depth suggests that attention to the mutual-cooperation payoff is systematically associated with the model's cooperative choices: games in which this payoff is more salient tend to yield higher cooperation rates. The sharp early negative dip in T-attention is consistent with the model rapidly identifying high T as the signature of games where defection is more attractive, effectively using T-salience as an implicit game-type classifier in early processing. These two signals shift across layers, with T dominating early and R dominating in mid-network, though they overlap rather than forming cleanly sequential stages. Rather than processing each payoff in isolation, Llama appears to represent their relative relationship.

The other two payoffs play different roles. The persistent negative S signal throughout all layers is consistent with attention to the sucker's payoff functioning as a defection-risk signal: regardless of layer, games in which the model allocates more attention to S are games where it cooperates less. P shows no consistent signal across layers; unlike S and T, P does not serve as a contextually salient cue that differentiates the model's cooperative behavior across the game space. This is especially informative because both R and P are fixed across all games. The positive R signal therefore cannot reflect the model responding to a higher R, since R never changes.

Equivalent plots for Qwen and Mistral are provided in Figure \ref{fig:attention_othermodels}; neither model replicates this pattern, with Qwen showing no sustained signal for any payoff and Mistral showing only a weak, noisy echo of the R-positive correlation. Thus, the model that best reproduces human cooperation patterns is also the model with the clearest layer-dependent organization of payoff salience. All of these results remain correlational: they show that attention to payoff values is associated with cooperation decisions, but they do not establish that attention causally determines those decisions.

\subsection{Extending the experimental grid: novel in-silico games\label{subsec:novelgames}}

\begin{figure}[t!]
    \centering    \includegraphics[width=\linewidth]{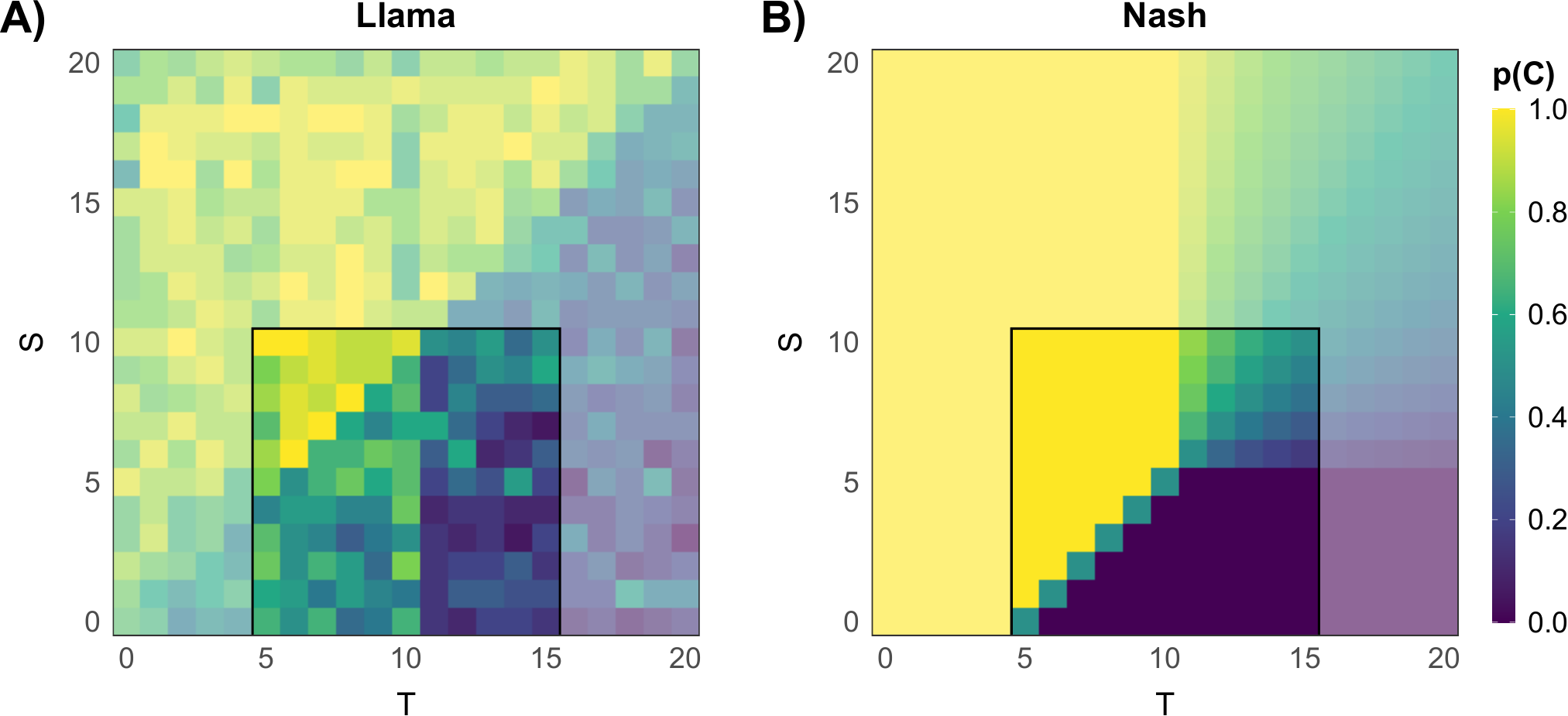}
    \caption{\textbf{Cooperation matrices of Llama and the Nash equilibrium, extended through simulations.} Each tile represents a game defined by the parameter combination of $(S,T)$, with average cooperation ranging from $0$ (purple: no cooperation) to $1$ (yellow: full cooperation). Panel A: Cooperation matrix of Llama. The non-transparent region delineated by a black border shows simulations with parameter combinations that were already part of the original human experiments. The transparent region shows combinations of $S$ and $T$ that were not played by the human participants. Panel B: Matrix showing the rational choices predicted by Nash equilibrium (see Methods Section \ref{subsec:analytic}) covering both original (non-transparent) and novel games (transparent). Simulations across the extended parameter grid generate testable hypotheses that can be validated through preregistered experiments to determine which extrapolation method for novel games better aligns with empirical human behavioral patterns.}
    \label{fig:novel_games}
\end{figure}

Using the exact same experimental setup, we have Llama play games with parameter combinations outside the regions covered in the original experiments with human participants. Panel A of Figure \ref{fig:novel_games} shows this new, extended cooperation matrix with novel combinations of $S$ and $T$ (shown as transparent regions) alongside simulated games already played in the experiments (non-transparent, delineated with a black border).

Llama exhibits high cooperation when $S \ge T$, extending the diagonal $S = T$ pattern into the enlarged cooperation matrix and demonstrating the importance of this parameter condition. Within the region where $S \ge T$, there is no apparent effect of the values $P$ and $R$ on the tendency to cooperate. However, when $S$ and $T$ have values near 0, the separation becomes less distinct, displaying a random-like cooperation pattern when $S < 5$ and $T < 5$. In the other half of the matrix, where $S < T$, the value of $R$ exerts a clear influence: there is a marked difference between regions where $T > R$ (exhibiting lower cooperation) and $T \le R$ (higher cooperation). Examining horizontal cross-sections at constant $S$ reveals that increasing $T$ reduces cooperation, though not linearly. Similarly, vertical cross-sections at constant $T$ show that increasing $S$ increases cooperation. We repeated the same extended-grid simulation for Qwen and Mistral. Results are shown in Figure~\ref{fig:extendedgames_othermodels} and discussed in Subsection~\ref{app:subsec:extension_othermodels}.

If we were instead to use Nash equilibrium solutions to predict the novel games (Panel B of Figure \ref{fig:novel_games}), we would find two pure regions of total defection and total cooperation separated by a boundary where the average cooperation is 0.5, along with a third region characterized by mixed equilibrium. Both the Llama and Nash simulations provide empirically testable hypotheses about human behavior in unexplored parameter regions. To decisively evaluate which set of predictions, based on classical rationality assumptions or LLM-based modeling, better anticipates actual human behavior, we have preregistered\footnote{\href{https://aspredicted.org/fe6z2k.pdf}{https://aspredicted.org/fe6z2k.pdf}\label{fn:prereg}} the fully-specified experimental setup that will allow future experiments to establish which approach aligns more closely with actual human behavior in these novel games.

\section{Discussion}

In this study, we evaluated LLMs as behavioral imitators of humans by replicating game-theoretic experiments. To this end, we developed a systematic answer extraction mechanism that turned the initial near-random LLM responses, obtained when prompting for simple direct answers, into meaningful cooperation patterns. Of the three evaluated models, Llama reproduced aggregate human cooperation most faithfully, and our attention-based analyses suggest this closer behavioral alignment is accompanied by a structured, layer-dependent organization of payoff salience. Extending the game parameter space beyond the original human experiments, we further derived testable predictions for novel strategic settings.

\subsubsection*{Importance of robust prompting setups}

Simple extraction produces nearly random responses. When models are asked to provide a one-word answer, they seem to default to surface-level pattern matching. By requesting detailed explanations (Double Extraction), we leverage the phenomenon whereby models perform better when allowed to "think aloud" by articulating their reasoning to improve their logical consistency \cite{wei2022chain}. The Multi-step Prompt further decomposes complex reasoning into manageable steps, reducing complexity and thereby minimizing errors. Finally, the Logical Verifier acts as a quality control mechanism, filtering out responses containing mathematical errors or logical contradictions. This layered approach functions as a form of "attention check" for LLMs, ensuring that recorded decisions exhibit systematic behavioral patterns consistent with strategic decision-making rather than random or inconsistent outputs. Together, these techniques help LLMs provide the consistency and robustness needed for social simulation.

We show that, in contrast to earlier work\cite{sreedharSimulatingHumanStrategic2024}, persona-based prompting is not necessary to replicate aggregate behavioral patterns. Rather than attempting to simulate individual participants, focusing on population-level patterns provides a more tractable and robust approach. This finding simplifies the simulation process while maintaining predictive validity for the collective behavioral phenomena of interest. This simplification should not be read as a dismissal of the micro level. Accurately simulating individual decision-makers remains an open and substantial challenge. We see this as an important direction for future work, one that would extend LLM-based behavioral modeling from population-level replication towards individual-level prediction.

\subsubsection*{Behavioral phenotyping of LLMs}

The phenotyping results reinforce the picture emerging from the cooperation matrices. The shared envious profile of humans and Llama suggests that the dominant heuristic in both is a comparison of one's own payoff across the two choices relative to the opponent's, rather than minimax reasoning or unconditional cooperation. This is a cognitively plausible and computationally simple decision rule, consistent with a bounded rationality interpretation, which is discussed below. Qwen's flat phenotype profile and Nash-like cooperation matrix, suggests that it does not settle into any single human-recognizable heuristic but instead approaches something closer to game-theoretically optimal reasoning, averaging across strategies in a way that resembles no particular archetype. Mistral's high Trustful and Optimist scores reflect its overall elevated cooperation rate rather than a principled always-cooperate strategy, and its weak Pessimist signal is consistent with the vertical cooperation boundary observed in Figure \ref{fig:refinement_models}, which tracks R rather than P. This characterization should be read alongside the substantially higher rate of unverified responses for Mistral noted in Subsection~\ref{subsec:problemgames}: roughly half of its recorded choices bypass logical verification, so part of its elevated and noisier cooperation pattern may reflect logical inconsistency rather than a genuine behavioral tendency.

An intriguing anomaly appears in Llama's behavior at the extreme parameter values near (0,0), where cooperation appears random-like. This may reflect numerical instability when payoffs approach zero, making the strategic structure of the game less salient to the model. Further investigation of edge cases in the parameter space could reveal the boundaries of LLMs' strategic reasoning capabilities.

\subsubsection*{Mechanistic Interpretability}

A key advantage of using open LLMs is that their behavior can be studied not only at the level of inputs and outputs, but also through their internal representations. Our attention analysis illustrates how this can provide useful clues about the computations associated with model decisions. In Llama, payoff salience was organized in a structured, layer-dependent way: attention to the mutual-cooperation payoff was positively associated with cooperation, while attention to the temptation payoff was negatively associated with cooperation in early layers. Rather than treating each payoff as an isolated number, Llama appears to encode relationships between payoffs, especially the contrast between the reward for mutual cooperation and the temptation to defect. This resembles the kind of comparison a boundedly rational agent might use when deciding whether cooperation is worthwhile, and may help explain why Llama reproduces human cooperation patterns more closely than Qwen or Mistral.

However, we stress that this analysis is correlational: we cannot distinguish between attention causally driving cooperation and both being jointly determined by the underlying game structure. The structured saliency profile of Llama, absent in Qwen and Mistral, may be part of what makes the model the closest behavioral approximation to human players in our data, though this should not be taken to imply that human payoff processing follows a similar organization: similar behavior can clearly arise from very different underlying computations.

At the same time, this analysis remains only an initial step toward mechanistic understanding. The patterns we identify are aggregated across games and attention heads, per layer. They show that Llama's behavioral alignment with humans is accompanied by a structured internal salience profile, but they do not yet reveal the full computation by which the model transforms payoff information into decisions. Future work should therefore combine behavioral replication with more fine-grained interpretability methods to determine whether the latent heuristics suggested by these aggregate patterns correspond to identifiable mechanisms inside the model.

\subsubsection*{AI for science}

Our extension of the experimental parameter space, while modest compared to all logically possible combinations, was strategically designed to systematically explore regions adjacent to the original experimental grid and to make it symmetric. This approach allows us to test whether patterns observed in the original games generalize smoothly to nearby parameter values or whether discontinuities emerge at certain boundaries. The original games represent only a small subset of the possible experimental space, and our extensions provide a principled way to extrapolate beyond known regions while maintaining interpretability.

A key distinction emerges between prediction using the analytical Nash equilibrium solution versus the LLM-based approach. The analytical solution predicts regions with intermediate cooperation rates (around 0.5) corresponding to mixed equilibria, particularly visible in the Snowdrift Game region of Panel B in Figure \ref{fig:novel_games}. These smooth gradients do not appear in Llama's behavior (Panel A), which instead exhibits overall more discrete, binary-like cooperation patterns. Mixed equilibria would require players to randomize their strategies in precise proportions, which is a sophisticated form of strategic reasoning that can be considered psychologically implausible for most players. Instead, both humans and LLMs likely tend to employ simpler decision heuristics based on comparing payoffs or identifying dominant features of the game structure. The closer alignment of Llama with human patterns rather than Nash predictions suggests that LLMs may capture the psychological and heuristic processes underlying human decision-making in ways that purely analytical Nash equilibrium solutions cannot. 

\subsubsection*{Implications for LLMs as behavioral imitators\label{subsubsec:implication}}

Our findings are consistent with the theory of bounded rationality, which emphasizes that models of social behavior must account for the limited access to information and, crucially, the computational capacities that are actually possessed by people \cite{simonBehavioralModelRational1955}. In cognitive science, a simplicity principle highlights our tendency to adopt simple strategies when faced with complex tasks \cite{feldmanSimplicityPrinciplePerception2016}. Empirically, researchers have found that we follow simple heuristics when choosing social connections \cite{martin-gutierrezHiddenArchitectureConnections2024} and that limited cognitive capacities explain the structure of our social circles \cite{tamaritCognitiveResourceAllocation2018}. However, LLMs' human-like behavior raises a different question. These games are mathematically simple, and sufficiently expressive models should in principle be able to identify the payoff-maximizing response when explicitly asked to choose the option that yields the most points. There is therefore no obvious computational reason for an LLM to depart from fully rational play, or to behave as if it were relying on simple heuristics. Although fully addressing this question falls outside the scope of the paper, we can  offer a tentative hypothesis: through training and alignment, some LLMs (in our case, Llama) become imitators of human behavior \cite{benderDangersStochasticParrots2021}, so rather than choosing optimal strategies, they tend to behave the way humans typically would.

The contrast between analytical rational-choice models and LLM-based behavioral models extends beyond predictive performance. Classical rational models, such as those based on Nash equilibrium, provide a clear theoretical rationale for their predictions: they specify how choices should follow from incentives, equilibria, and assumptions about utility. Our results should not be read as a rejection of such models. Their strength lies precisely in making assumptions explicit, allowing researchers to explore how different utility formulations or rationality assumptions lead to different behavioral predictions. LLM-based models are useful in a different way. Rather than representing a single formal decision rule, they may approximate the heterogeneous mixture of tendencies that shapes aggregate human behavior. But they do so as black boxes whose internal reasoning remains inaccessible to a substantial degree. Nevertheless, unlike traditional models, LLMs can be interrogated through prompting and probed through analyses of their internal representations, offering a route to uncovering the latent heuristics that drive their choices. In this sense, LLMs complement rather than replace rational models. A deeper synthesis between interpretability and predictive validity may ultimately lead to a more complete understanding of social decision-making.

The distinct cooperation profiles across models provide additional insights. Qwen's behavior closely aligns with the Nash equilibrium (Table \ref{combined_table}), which may reflect either its exposure to more game-theoretic literature during training or superior logical reasoning about optimal strategies. Importantly, we deliberately avoided using game theory terminology such as "cooperate" and "defect" in our prompts to prevent triggering memorized responses, yet Qwen still converges toward rational play. While this makes Qwen an excellent test of strategic reasoning, it may render it less useful for predicting human behavior in novel games, since humans systematically deviate from Nash equilibrium predictions (see Panel A of Figure \ref{fig:refinement_models}). Mistral exhibits an intermediate pattern, showing sensitivity to the relative payoff structure but not fully optimizing in the game-theoretic sense.

To address the concern that LLMs might merely reproduce memorized information about the original experiments\cite{barrieEmergentLLMBehaviors2025}, we note several points. We acknowledge that we cannot rule out that some parameter combinations covered in our extended grid appeared in training corpora through other experimental studies. However, memorization from prior work would not be straightforward. As shown in Table~\ref{tab:prior_studies}, existing game-theoretic experiments use widely varying payoff scales, experimental contexts, and operationalizations: payoff values range across several orders of magnitude, studies mix one-shot and repeated games, and cooperation rates vary substantially even for structurally similar parameter combinations. Translating any memorized pattern from these heterogeneous sources into our specific normalized payoff grid (R = 10, P = 5) would require non-trivial abstraction that goes well beyond direct retrieval. Beyond the heterogeneity in payoff structures and experimental contexts, data availability varies considerably across these studies. Several are published as books or in paywalled journals without open data repositories, making it unlikely that quantitative results from all those sources were present in model training corpora. Even for openly available studies, cooperation rates are typically reported in manuscript tables or figures rather than structured datasets, further complicating any direct retrieval mechanism.

Most importantly, if direct memorization of experimental results were the primary mechanism, models trained on broadly similar internet-scale corpora and subjected to identical prompting would be expected to converge to similar behavioral profiles. Instead, we observe markedly distinct cooperation patterns across Llama, Qwen, and Mistral, strongly suggesting that the differences reflect genuine behavioral variation rather than direct retrieval of stored results. We acknowledge this as a limitation and encourage future work to probe training data access more directly.

\subsubsection*{Closing the loop: future human experiments and theory-building}

The validity of our approach can only be definitively assessed through future experiments with human participants. While the precise mechanisms that enable LLMs to capture patterns of human experience and behavior may remain opaque and debated\cite{larooijLargeLanguageModels2025}, we can evaluate the models' predictive power in a straightforward way: by carrying out subsequent experiments with humans, ideally following strictly specified experimental setups. To enable unambiguous evaluation, we have publicly pre-registered\footref{fn:prereg} the complete experimental design that needs to be conducted with human participants to validate our extended simulation results. This pre-registration specifies all details of the experimental protocol, preventing post-hoc adjustments and ensuring rigorous hypothesis testing.

More broadly, our work exemplifies the potential of AI for scientific discovery in the social and behavioral sciences. By using LLMs as computational models of human behavior, we can rapidly explore vast experimental spaces that would be prohibitively expensive or otherwise too burdensome to investigate with human participants alone. This approach follows a virtuous cycle: validate models on existing human data, use models to generate novel hypotheses, then conduct targeted human experiments to test those hypotheses. The transparent, pre-registered nature of this process ensures that AI-generated predictions face genuine empirical tests, maintaining the rigor of scientific inquiry while dramatically expanding its scope and efficiency. As LLMs continue to improve, this methodology can be expected to become a standard tool for behavioral and social scientists, enabling systematic in-silico exploration of experimental landscapes before committing resources to the actual experiments involving humans.

\section{Methods}

\subsection{Experimental Design}

As in the original experiments, we keep $R$ and $P$ constant at 10 and 5, with $S$ ranging from 0 to 10 and $T$ ranging from 5 to 15. To extend the parameter space, we varied $S$ and $T$ more broadly: $S$ ranges from 0 to 20, with 21 possible values, and $T$ ranges from 0 to 20, with 21 possible values. In total, we simulate 441 games per model, 320 more than in the original experiments. Each one-shot and anonymous game was played 20 times to account for inherent stochasticity. Our main objective was to construct a cooperation matrix in which each element represents a specific game defined by its $(S,T)$ parameter combination, with the element's value indicating the average cooperation rate for that game. For each simulated game round, we record the model's choice as $1$ if it chose to cooperate (choice A or B, depending on random assignment) and $0$ if it chose to defect. After completing all 20 iterations of each game, we calculate the average cooperation rate per game by averaging these binary outcomes. We compare the cooperation matrices from the original 121 games against the human and Nash benchmarks using Pearson's correlation and mean squared deviation. We quantify distributional differences in per-game cooperation rates for these same 121 games using the Wasserstein-1 distance (Earth Mover's Distance), which measures the minimum cost of transforming one empirical distribution into another on the unit interval. 

\subsection{Prompt engineering and model parameters\label{subsec:modelparam}}
Model prompts consist of three different components (for the full text of prompts see Subsection~\ref{app:subsec:prompts} in the Appendix):

\begin{itemize}
    \item \textbf{System prompt.} Specifies the role assigned to the model. We use the default "You are a helpful assistant.".
    \item \textbf{User prompt.} Contains the request made to the model. This prompt includes the game instructions and requests a choice.
    \item \textbf{Assistant prompt.} Contains the model's response. This component can be used to simulate a conversation history of previous exchanges. In practice, the verifier and extractor stages (see Methods Section \ref{subsec:method_extract}) instead embed the tested model's response as quoted text within a subsequent user prompt, rather than as a true assistant turn.
\end{itemize}

All simulation code\footnote{\href{https://github.com/maxpel/llmgames}{https://github.com/maxpel/llmgames}} is written in Python, using the vLLM library for efficient inference. Since each model requires a different prompting format, we develop functions to adapt the general prompt structure to the format required by each LLM while keeping the text content of the prompt parts identical. We load all models in 16-bit floating-point precision (\texttt{dtype\,=\,float16}) with a maximum 
context length of $8{,}192$ tokens, a GPU memory utilization ceiling of $0.44$, and tensor parallelism disabled 
(\texttt{tensor\_parallel\_size\,=\,1}). For response generation in the main results we use a temperature of $\tau = 0.8$ and a maximum of $1{,}000$ output tokens; for the logical verifier and answer extractor (Qwen) we use $\tau = 0.3$ and a maximum of $50$ output tokens. We assess robustness to this choice in Appendix~\ref{app:varianceviolins}, where we additionally vary $\tau \in \{0.0, 0.1, \dots, 1.0\}$ for response generation. Figure~\ref{fig:tempperformance} shows that our results are robust to temperature variation; temperature $0$ generally leads to a drop in performance. For the attention-based payoff saliency analysis (Section~\ref{subsec:attention_analysis}), since vLLM does not expose attention weights, we instead load models directly via the Hugging Face \texttt{transformers} library, with eager attention computation (\texttt{attn\_implementation="eager"}) and the same 16-bit floating-point precision.

We conduct all experiments on MareNostrum~5 at the Barcelona 
Supercomputing Center (BSC). All jobs run on one node using 20 Intel Sapphire Rapids 8460Y+ CPU cores, 1 H100 GPU with 
64\,GB HBM2e memory, and up to 122\,GB RAM. The main 121-game simulations are drawn from the temperature ablation runs (Appendix~\ref{fig:both}), which comprised 33 array tasks ($3~\text{models} \times 11~\text{temperatures}$); wall-clock time at $\tau = 0.8$ was 19, 22, and 33 minutes for Llama, Qwen, and Mistral respectively, with the full ablation totalling approximately 16.4~GPU-hours. The extended 441-game simulations (Subsection~\ref{subsec:novelgames}) comprised 3 additional array tasks (one per model), completing in 38, 40, and 60 minutes for Llama, Qwen, and Mistral respectively ($\approx$2.3~GPU-hours total).

\subsection{Adapting the original game instructions to LLMs}
\label{subsec:method_adapt}

We initially attempted to directly use the original instructions from the human experiment with minimal modification. After running initial test simulations with the original game instructions, we observed that the resulting prompts were overly verbose and that models prioritized generating discursive, open-ended responses rather than engaging in strategic reasoning about payoff structures. We therefore conducted an iterative refinement process, arriving at the following key modifications:

\begin{itemize}
    \item \textbf{Removal of non-essential context.} We removed all contextual information about the experimental motivation, participating institutions, and background details. We retained only the core rules of the game and the payoff structure.
    \item \textbf{Explicit statement of one-shot nature and simultaneity.} We explicitly stated that each game consisted of a single round with simultaneous decision-making to prevent an observed tendency for models to adopt cooperative strategies aimed at influencing future rounds or signaling intent to the other player.
    \item \textbf{Avoidance of explicit game-theoretic terminology.} We observed that using game theory vocabulary such as "opponent", "cooperate" or "strategies" led models to activate memorized game-theoretic scenarios rather than reasoning about the specific payoff structure (see also the discussion of memorization in Section~\ref{subsubsec:implication}). We therefore adopted neutral vocabulary such as "other player" instead of "opponent" and replaced "C" and "D" with "A" and "B" as the two decision options, whose label-to-strategy assignment we additionally randomized across games.
    \item \textbf{Clarification that the game has no competitive objective.} We observed that models often treated the games as zero-sum competitions, choosing to defect to "beat" the other player even in Harmony Games where defection is the strictly dominated strategy and thus disadvantageous for the defecting player regardless of the opponent's choice. We therefore explicitly stated that each player's prize depends solely on their own points earned, not on relative performance or point differences.
    \item \textbf{Clarification of the prize structure.} We replaced the original lottery ticket rewards with a clearer monetary conversion: 10 euros per point. We also added a concrete example of the calculation (e.g., "if player 1 earns 9 points and player 2 earns 5 points, player 1 receives 90 euros and player 2 receives 50 euros") to ensure models understood the direct relationship between points and payoffs and to reduce irrelevant cross-player payoff comparisons and expected value calculations.
\end{itemize}

These changes were developed through manual prompt engineering informed by established best-practice guides \cite{fagbohun2024empirical, amatriain2024prompt}. Our iterative refinement process consisted of: (1) running simulations with the current prompt, (2) analyzing model responses across multiple games to identify systematic errors or misconceptions, (3) modifying the prompt to address these issues, and (4) validating that the problematic patterns were resolved in subsequent simulations. This cycle was repeated until the models answers were largely consistent with the game structure and payoff incentives, with remaining edge cases addressed through the extraction and verification pipeline described in Methods Section~\ref{subsec:method_extract}.

\subsection{\label{subsec:method_extract}Answer extraction approaches}

We develop a four-stage extraction methodology with progressively increasing complexity to reliably extract and validate strategic choices from model outputs.

\subsubsection{Generation of long answers}

Initially, we prompted models for brief responses that can be easily parsed using regular expressions. However, as we show in Figure \ref{fig:refinement_models} and \ref{fig:prompting_stages}, these short, simple and direct answers exhibited high variability and often appear arbitrary. In contrast, when we allow models to generate longer responses, they produce more logically coherent reasoning. This suggests that having the space to articulate explanations helps the models perform, a phenomenon observed in prior work showing that encouraging models to "think step-by-step" enhances task accuracy\cite{wei2022chain}.
Based on these observations, we adopt a Double Extraction approach: First, elicit a detailed response from the tested model using the instruction prompt and game-specific payoff values (Subsection~\ref{app:subsec:prompts} in the Appendix, with model parameters as described in Methods Section~\ref{subsec:modelparam}), then use a second LLM to extract the final choice from this explanation, building on best-practice approaches in survey response generation with LLMs\cite{ahnertSurveyResponseGeneration2025}.

\subsubsection{Extraction of short answers}

The second step of the Double Extraction process extracts the final choice from the detailed response. We provide the long answer generated by the tested LLM as input to a second LLM (Qwen), which we prompt to identify only the player's choice: A or B. A simple regular expression function then parses this output to extract the selected letter. If the extraction successfully identifies either A or B, we record the choice as valid; otherwise, we flag it as invalid and require the tested model to replay that game. We select Qwen for this extraction step to maintain consistency with its role in the logical verification stage (see Methods Subsection \ref{subsec:logicalverifier}), so that both quality-control components of the pipeline are governed by the same model (sampling parameters as described in Methods Section~\ref{subsec:modelparam}).

\subsubsection{Multi-step prompt}
Although the double extraction approach yielded improved results compared to simple extraction, manual inspection revealed that the models still produced frequent logical inconsistencies and factually incorrect statements. Previous research \cite{kojima2022large, wei2022chain} has shown how results can be improved by prompting models to articulate step-by-step reasoning. Breaking down complex reasoning tasks into manageable steps has also  been shown to be helpful \cite{fagbohun2024empirical,amatriain2024prompt}. We therefore design prompts that explicitly guide models to decompose their reasoning into discrete steps.

These steps need to be specific enough to scaffold the reasoning process, yet generic enough to avoid biasing models toward particular strategies. We iteratively test multiple multi-step prompt variants, each of which instructs models to: (1) group the four possible outcomes by their own choice (A or B), and (2) compare the payoffs within each group. To identify the optimal prompt, we evaluate performance in the $S \geq T$ region (Harmony Games), where cooperation is the only sensible choice, the same validation region used for the logical verifier prompt in Section~\ref{subsec:logicalverifier}. We select the prompt version that yields the highest average cooperation rates for both Llama and Mistral in this region. This multi-step instruction was inserted into the final prompt immediately following the description of game outcomes.

\subsubsection{Logical verifier}
\label{subsec:logicalverifier}

Despite the multi-step prompt improvements, manual inspection continued to reveal logical and mathematical errors in model responses. We therefore implement a logical verification layer as a final quality control mechanism.

We first conduct systematic error analysis, cataloging the most common mistakes produced by Llama and Mistral. Based on these patterns, we design a verification prompt that instructs Qwen (maintaining consistency with the extraction step) to classify each long-form response as "good" or "bad" based on its logical validity. The prompt consists of two components: First, we explicitly define criteria for valid and invalid responses in bullet-point format, emphasizing the specific logical errors we have identified (e.g. incorrect arithmetic comparisons, misunderstanding of outcome probabilities or inconsistencies between reasoning and final choice). Second, we incorporate diverse examples of both acceptable and flawed responses, employing the few-shot learning technique \cite{brown2020language}, which has been shown to enhance logical reasoning even in mid-sized models like those used in our experiments. From a pool of ten annotated examples, we randomly sample five without replacement for each verifier call, so that the prompt does not repeatedly expose the model to the same fixed set of examples. We test multiple versions of this combined prompt with varying example sets and evaluate their performance through simulation. Following the same validation approach used for the multi-step prompt, we select the version that maximizes cooperation rates in the $S \geq T$ region (Harmony Games).

\subsection{Conditions for answer validity}

Each game required a valid response to be included in the final dataset. Our validation pipeline proceeded as follows: (1) the tested model generated a long-form response, (2) the logical verifier (Qwen) classified it as valid or invalid, (3) if valid, the extractor (also Qwen) identified the final choice (A or B). Games producing invalid responses at either verification or extraction stages were replayed in subsequent rounds until a valid response was obtained.

However, preliminary testing revealed that certain game configurations consistently produced logically flawed responses, trapping the simulation in infinite loops. To address this issue, we implemented an adaptive relaxation mechanism: if the number of invalid games remained unchanged between two consecutive rounds, we temporarily disabled the logical verifier for those problematic games, allowing them to proceed directly to the extraction step. This pragmatic compromise ensured simulation completion while maintaining quality control for the majority of responses (for details see Subsection \ref{subsec:problemgames}).

\subsection{\label{subsec:analytic}Computation of Nash equilibrium}

To compute Nash equilibrium cooperation rates, we simulated replicator dynamics from evolutionary game theory. When strategies compete, the fraction of strategy $i$ evolves according to:
\begin{equation}
    \dot{x}_{i,t} = x_{i,t}\left( \pi_{i}(x_{t}) - \bar{\pi}(x_{t})\right)
    \label{timevolution}
\end{equation}
where $x_{i,t}$ represents the fraction of the strategy $i$ at time $t$, $\pi_{i}(x_{t})$ is its average payoff and $\bar{\pi}(x_{t})$ is the average payoff of all strategies at time t.

\vspace*{\baselineskip}
\noindent With only two strategies (\emph{Cooperate}--C and \emph{Defect}--D), we just need to track the cooperator fraction. Setting $\Delta{t}=1$ in Equation~\ref{timevolution}, we obtain:
\begin{equation}
    x_{t+1} = x_{t}(1 - x_{t})\left( \pi_{C}(x_{t}) - \pi_{D}(x_{t})\right) +  x_{t}
    \label{evolutionC}
\end{equation}
where $x_{t}$ is cooperator fraction at time $t$, $\pi_{C}(x_{t})$ and $\pi_{D}(x_{t})$ are
cooperator and defector average payoffs, respectively. These payoffs are:
\begin{align}
    \pi_{C} = x_{t}R + (1 - x_{t})S\\
    \pi_{D} = x_{t}T + (1 - x_{t})P
\end{align}
with $R = 10$ and $P = 5$ as in the original experiments\cite{poncela2016humans}, while $S$ and $T$ vary by game.

\vspace*{\baselineskip}
\noindent We started from the initial condition $x_0=0,5$ and iterated until $t=10^3$ or until one of four outcomes occurred (tolerance $\epsilon = 0.1$):
\begin{itemize}
    \item \textbf{Total Cooperation:} $x_{t} \ge 1 - \epsilon$ (Harmony Game).
    \item \textbf{Total Defection:} $x_{t} \le \epsilon$ (Prisoner's Dilemma).
    \item \textbf{Mixed Equilibrium:} $|\pi_C - \pi_D| \le \epsilon$, where payoffs equalize and cooperator fraction stabilizes between $0$ and $1$ (Snowdrift Game, Stag Hunt diagonal).
    \item \textbf{Periodicity:} Oscillation between two $x_{t}$ values; we recorded their average (some Snowdrift games).
\end{itemize}

\noindent We have verified these results by performing a fixed-point and stability analysis of the replicator dynamics corresponding to the $2\times2$ symmetric game defined by the payoff matrix:
\[
\begin{pmatrix}
R & S\\
T & P
\end{pmatrix}
\]
  
Let $x\in[0,1]$ denote the fraction of cooperators in the population.  
Using Equation~\ref{timevolution}, the replicator dynamics can be written as
\begin{equation}
\dot{x} = x(1-x)\big[\pi_C(x) - \pi_D(x)\big]
\label{replicator}
\end{equation}
where
\begin{align}
\pi_C(x) &= R x + S(1-x)\\
\pi_D(x) &= T x + P(1-x)
\end{align}
Equation~\ref{replicator} can be rewritten in the compact form
\begin{equation}
\dot{x} = x(1-x) g(x)
\label{replicator_g}
\end{equation}
where
\begin{equation}
g(x) = \pi_C(x) - \pi_D(x) = (R - T - S + P)\,x + (S - P)
\label{gx}
\end{equation}
The function $g(x)$ measures the instantaneous advantage of cooperation over defection:  
if $g(x)>0$, cooperation tends to increase; if $g(x)<0$, it tends to decrease.  
Fixed points correspond to values of $x$ for which $\dot{x}=0$, i.e.
\[
x\in\{0,1\}\quad \text{or}\quad g(x)=0.
\]

\paragraph{(i) Boundary fixed points.}

To test the stability of the homogeneous states $x=0$ (all defect) and $x=1$ (all cooperate), we can inspect the sign of $\dot{x}$ in their neighborhoods:

\begin{itemize}
    \item Near $x=0$ (almost all defect), $x$ is small and the factor $(1-x)\approx1$, so
    \[
    \dot{x}\approx x\, g(0),
    \quad\text{where}\quad g(0)=S-P.
    \]
    If $g(0)>0$ ($S>P$), then $\dot{x}>0$ for small $x$: cooperation increases and $x=0$ is \emph{unstable}.  
    Conversely, if $g(0)<0$ ($S<P$), then $\dot{x}<0$ near $x=0$ and the population returns to full defection: $x=0$ is \emph{stable}.
    
    \item Near $x=1$ (almost all cooperate), set $x = 1 - \varepsilon$ with $\varepsilon \ll 1$. 
    Then
    \[
    \dot{x} \approx \varepsilon\, g(1),
    \quad\text{where}\quad g(1)=R-T.
    \]
    If $g(1)>0$ ($R>T$), then $\dot{x}>0$ for $x<1$, so the system moves back toward $x=1$: cooperation is \emph{stable}.
    If $g(1)<0$ ($R<T$), then $\dot{x}<0$ and $x$ decreases away from $1$: cooperation is \emph{unstable}.
\end{itemize}

\noindent These two simple conditions,
\[
x=0\ \text{defection stable if } S<P,\qquad
x=1\ \text{cooperation stable if } R>T,
\]
partition the parameter space into regions of dominance and coordination.

\paragraph{(ii) Interior fixed point.}

An interior equilibrium $x^*$ exists whenever $g(x^*)=0$, that is,
\begin{equation}
x^* = \frac{P - S}{R - T - S + P}.
\label{xstar_explicit}
\end{equation}
The interior fixed point exists if ($0\leq x^* \leq1$) and $(R - T - S + P)\neq0$. Let us rewrite its expression as

\[
x^*=\frac{P-S}{D},\qquad D:=R-T-S+P.
\]
Then
\[
x^*>0 \;\Longleftrightarrow\; (P-S)\,D>0
\quad\text{and}\quad
x^*<1 \;\Longleftrightarrow\; \frac{R-T}{D}>0 \;\Longleftrightarrow\; (R-T)\,D>0,
\]
since \(1-x^*=\dfrac{R-T}{D}\).  
Therefore,
\[
0<x^*<1 \;\Longleftrightarrow\; (P-S)\,D>0 \;\text{ and }\; (R-T)\,D>0,
\]
i.e., \(D\) must have the \emph{same sign} as both \(P-S\) and \(R-T\). Therefore, we have two regions of existence:
\[
\begin{cases}
D>0 \;\Rightarrow\; S<P \ \text{and}\ T<R \quad  \\
D<0 \;\Rightarrow\; S>P \ \text{and}\ T>R 
\end{cases}
\]

\noindent\textit{Specialization to our parameters.}
With \(R=10\) and \(P=5\),
\[
0<x^*<1 \;\Longleftrightarrow\;
\big(S<5 \ \text{and}\ T<10\big)\ \ \text{or}\ \ \big(S>5 \ \text{and}\ T>10\big),
\]
corresponding exactly to the bottom-left and top-right regions of the \((T,S)\) plane.

\vspace*{\baselineskip}

\noindent To test its stability, we can linearize Equation~\ref{replicator_g} around $x^*$:
\[
\dot{x} \approx x^*(1-x^*)\, g'(x^*)\, (x - x^*),
\]
since $g(x^*)=0$.  
The factor $x^*(1-x^*)>0$, so the sign of $g'(x^*)$ determines the stability: Stable if $g'(x^*)<0$, unstable if $g'(x^*)>0$. From Equation~\ref{gx},
\[
g'(x) = R - T - S + P = D,
\]
which is constant. Therefore $x^*$ is stable if  $D=R - T - S + P < 0$, which corresponds to the top right quadrant of our region of study $(S>5 \ \text{and}\ T>10)$.

\vspace*{\baselineskip}
\noindent Intuitively, this condition means that when the gain from mutual cooperation ($R-P$) is smaller than the temptation and sucker effects combined ($T-S$), the dynamics settle into a stable coexistence between cooperators and defectors (anti-coordination).  
Otherwise, the dynamics are bistable, typical of coordination games.

\paragraph{(iii) Summary of Stability Criteria.}

\begin{center}
\begin{tabular}{lll}
\toprule
Condition & Stable point(s) & Game type \\
\midrule
$S<P$, $T>R$
(bottom right quadrant)& $x=0$ & Defection dominates (Prisoner’s Dilemma)\\
$S>P$, $T<R$
(top left quadrant)& $x=1$ & Cooperation dominates (Harmony)\\
$S<P$, $R>T$, $R-T-S+P>0$
(bottom left quadrant)& $x=0,1$ (bistable) & Coordination (Stag Hunt)\\
$S>P$, $R<T$, $R-T-S+P<0$
(top right quadrant)& $x^*$ (mixed) & Anti-coordination (Snowdrift)\\
\bottomrule
\end{tabular}
\end{center}

In the coordination (bistable) region, corresponding to $T<10$ and $S<5$, both $x=0$ (full defection) and $x=1$ (full cooperation) are locally stable, while the interior fixed point $x^*$ is unstable. Since in our simulations we set $x(0)=0.5$, the long-term outcome depends on the position of this initial condition relative to $x^*$:
\[
\begin{cases}
x(0)>x^* \ \Rightarrow\  \text{the population converges to full cooperation }(x=1),\\[4pt]
x(0)<x^* \ \Rightarrow\  \text{the population converges to full defection }(x=0).
\end{cases}
\]
Using Equation~\eqref{xstar_explicit}, this boundary between the basins of attraction is given by
\[
x^* = \frac{5-S}{15 - T - S},
\]
so that for $x(0)=0.5$ cooperation prevails whenever $T < 5 + S$, and defection otherwise. This condition precisely separates the two attraction basins observed in the numerical simulations.

\subsection{Behavioral Phenotyping}
\label{subsec:phenotyping_methods}

We characterize the behavioral profile of each model by comparing its cooperation matrix against four ideal phenotype matrices derived from the decision rules described in Poncela-Casasnovas et al.\cite{poncela2016humans}. Each ideal matrix is constructed by applying the corresponding decision rule across all 121 $(S,T)$ combinations: the Optimist cooperates if $T < R$, the Pessimist if $S > P$, and the Envious if $S \geq T$; the Trustful always cooperates, yielding a constant matrix of ones. Following the boundary convention used in the original study (see Figure~3 therein), we code the exact tie cases ($T = R$ for the Optimist, $S = P$ for the Pessimist) as a mixed value of $0.5$ rather than as a strict $0$ or $1$, since these boundary games do not have a determinate prediction under either decision rule. We compute the Pearson correlation between each model's empirical cooperation matrix and each ideal matrix across all 121 games. Because the Trustful matrix is constant, its correlation with any empirical matrix is undefined; we instead report the mean cooperation rate as the most informative proxy. We omit the Undefined phenotype from the original paper, which applied only to the subset of participants whose cooperation was close to random across games.

To visualize the overall similarity structure among models, humans, Nash equilibrium predictions, and the four ideal phenotypes (Figure~\ref{fig:ideal_phenotype_heatmap_mds}, Panel C), we additionally compute pairwise distances $d_{ij} = 1 - r_{ij}$ between all nine entities (Human, Llama, Qwen, Mistral, Nash, and the four ideal phenotype matrices), where $r_{ij}$ is the Pearson correlation between their respective cooperation vectors across all 121 games, using the same Trustful proxy substitution described above wherever one of the two vectors is constant. We embed the resulting $9 \times 9$ distance matrix in two dimensions using classical (metric) multidimensional scaling (\texttt{cmdscale} in R), and report the percentage of variance explained by each dimension from the corresponding eigenvalues. As a fit diagnostic for this two-dimensional embedding, we additionally report Kruskal's Stress Formula~1, computed post hoc as

\begin{equation*}
\text{Stress-1} = \sqrt{\frac{\sum_{i<j} \left(d_{ij} - \hat{d}_{ij}\right)^2}{\sum_{i<j} d_{ij}^2}},
\end{equation*}

where $\hat{d}_{ij}$ denotes the Euclidean distance between the corresponding pair of points in the two-dimensional configuration.

\subsection{Attention analysis to establish payoff saliency}
\label{subsec:attention_analysis}

To probe which payoff information the model attends to when forming cooperative decisions, we extract layer-wise attention weights from models across all transformer layers, using the Hugging Face \texttt{transformers}-based configuration described in Section~\ref{subsec:modelparam} rather than the vLLM pipeline used for the cooperation-rate simulations. For each of the 121 $(S,T)$ game configurations, we record the attention allocated from the final prompt token to the token(s) corresponding to each payoff value ($R$, $S$, $T$, $P$) in the payoff table presented in the instructions (see Subsection~\ref{app:subsec:prompts}), averaging over all attention heads within each layer. Payoff token positions are located via character-level offset mapping from the prompt text to tokenizer output; for payoff values that span more than one token (e.g. two-digit numbers under some tokenizers), attention is averaged across all covering tokens. For Llama, this yields a $121 \times 32$ matrix of attention weights per payoff. Because each game is independently presented with one of two random label-to-strategy orderings (Subsection~\ref{subsec:method_adapt}, bullet point on label randomization), we compute attention separately under both orderings for every game and average the two before correlating with cooperation rate, so that the resulting profile is not confounded by which letter was assigned to which strategy. We then compute Pearson correlation across the 121 games between these per-layer attention weights and the corresponding empirical cooperation rates, producing a correlation profile over network depth for each payoff. For easier visual comparison, we smooth profiles with a centred rolling mean of window size 3. Since $R$ and $P$ are held fixed across all game configurations ($R=10$, $P=5$), their attention correlations reflect contextual salience (how much the model revisits those tokens given the current $S$ and $T$) rather than sensitivity to value variation. We note that this analysis is correlational: we cannot distinguish between attention causally driving cooperation and both reflecting an underlying representation of game structure\cite{jain-wallace-2019-attention,wiegreffe-pinter-2019-attention}.

\bibliography{sample}

\section*{Acknowledgments}

ASC thanks the office of the vice-rector for research and innovation of the Universitat Oberta de Catalunya for granting her a one-year teaching leave to carry out research. This publication is paid by the BSC Computational Social Sciences and Humanities (CSSH) Laboratory, which is supported by the Generalitat de Catalunya.

\section*{Author contributions statement}

ACP: Conceptualization, Methodology, Formal Analysis, Data Curation, Visualization, Writing - Original Draft, Writing - Review and Editing. SMG: Original Conceptualization, Visualization, Formal Analysis, Writing - Review and Editing. ASC: Conceptualization, Writing - Review and Editing. MP: Original Conceptualization, Methodology, Formal Analysis, Data Curation, Visualization, Writing - Original Draft, Writing - Review and Editing. All of the authors approved the final manuscript for submission.

\section*{Additional information}

Code and data to fully replicate our analyses are available at 
\href{https://github.com/maxpel/llmgames}{https://github.com/maxpel/llmgames}. Human participant data from the original experiments\cite{poncela2016humans} is available at \href{https://zenodo.org/records/1127154}{https://zenodo.org/records/1127154}.

\clearpage
\appendix

\renewcommand{\thefigure}{A\arabic{figure}}
\setcounter{figure}{0}

\renewcommand{\thetable}{A\arabic{table}}
\setcounter{table}{0}

\section{Appendix}

\let\oldsubsection\subsection
\renewcommand{\subsection}{\FloatBarrier\oldsubsection}

\subsection{\label{app:varianceviolins} Effect of temperature}

\begin{figure}[H]
  \begin{subfigure}[t]{0.48\textwidth}
  \vspace{0pt}
  \centering
    \includegraphics[width=\linewidth]{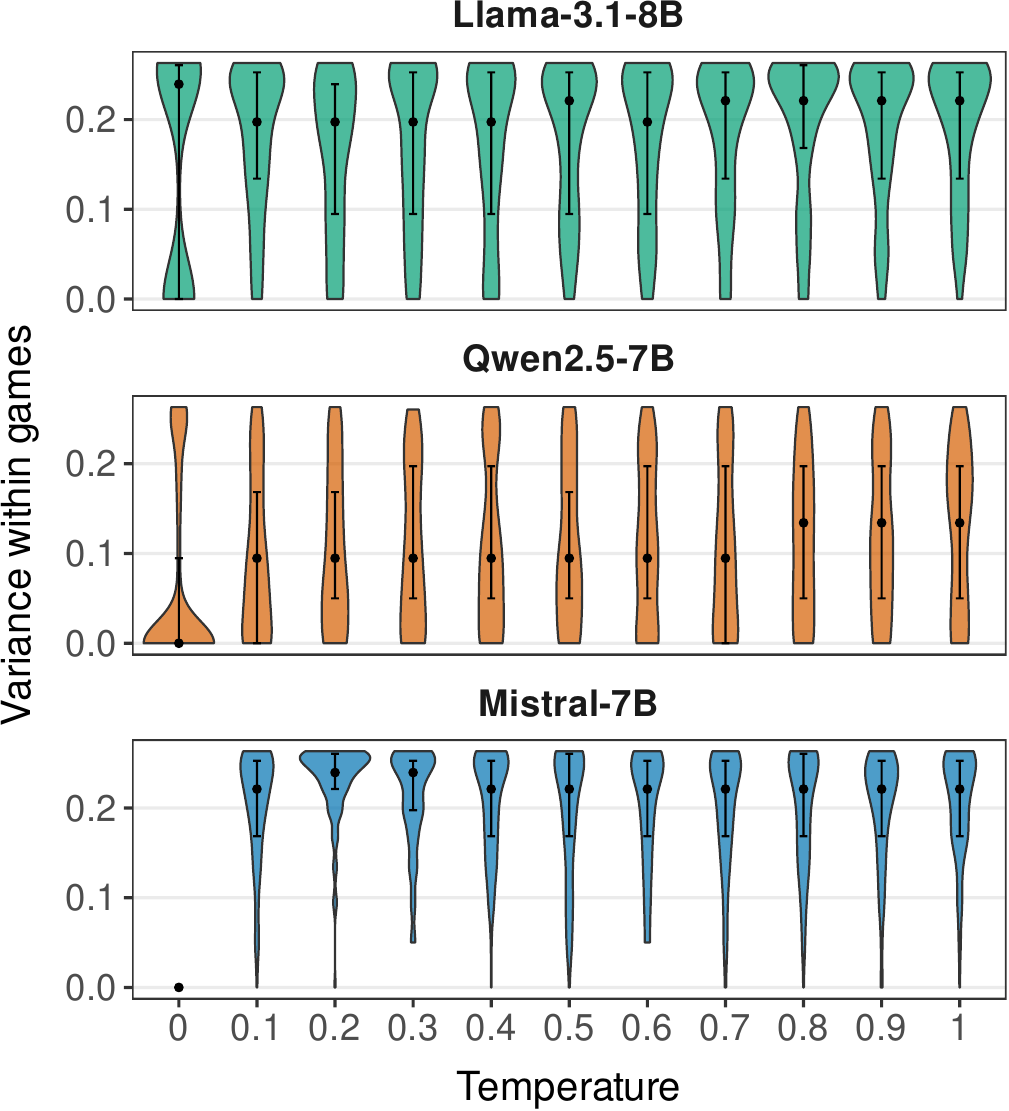}
    \caption{\textbf{Cooperation variance across games and model temperatures.} Distribution of variance across 121 games, each computed over 20 repetitions. Violin width shows density, error bars show IQR, points shows median. Increasing the tested model's temperature from 0 tends to increase within-game variance. Variance is not exactly zero at $\tau = 0$, since label-to-strategy randomization and the fixed $\tau = 0.3$ sampling used for the verifier and extractor (Qwen) remain active regardless of the tested model's temperature.}
    \label{fig:tempvariance}
  \end{subfigure}
  \hfill
  \begin{subfigure}[t]{0.48\textwidth}
  \vspace{0pt}
  \centering
    \includegraphics[width=\linewidth]{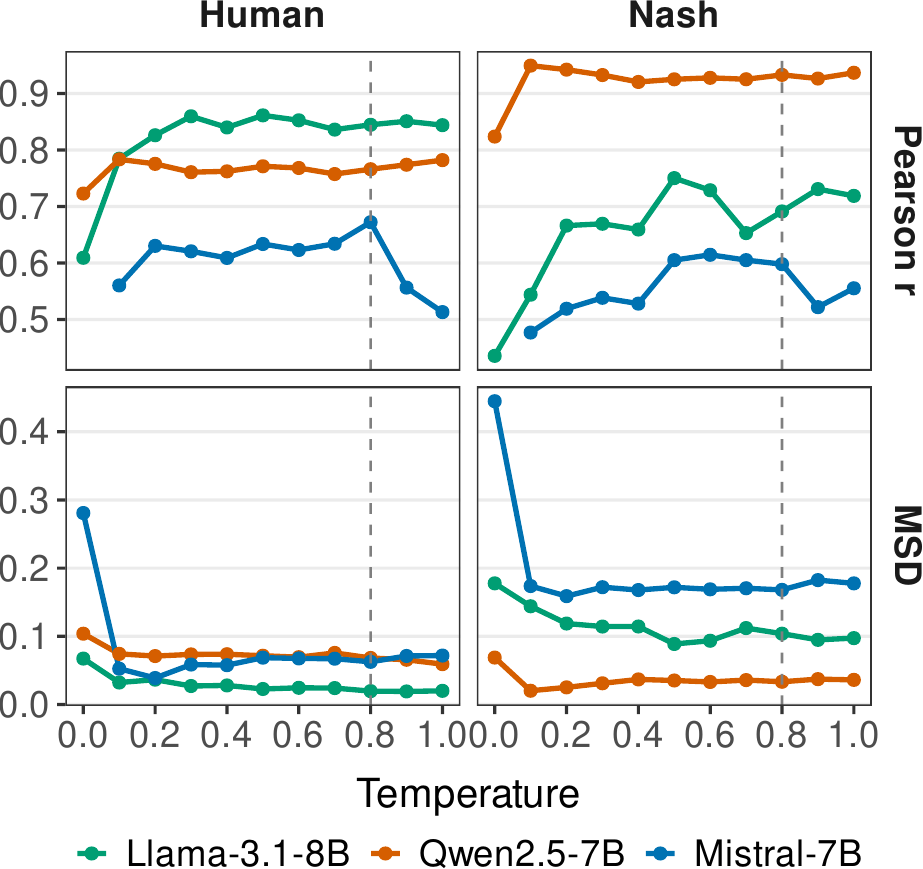}
    \caption{\textbf{Behavioral imitation across model temperature settings.} Pearson $r$ and MSD between model and human cooperation matrix. All models perform worse at temperature 0. Dashed line: temperature used for results in main text (0.8). Colors for models are the same as in Subfigure a).}
    \label{fig:tempperformance}
  \end{subfigure}
  \caption{\textbf{Effect of temperature parameter.} Higher temperature is introducing more variance within games. Except for very low temperature, results are generally robust to temperature changes.}
  \label{fig:both}
\end{figure}

We assess the robustness of our main results to the choice of sampling temperature by repeating the 121-game simulation at eleven temperature values, $\tau \in \{0.0, 0.1, \dots, 1.0\}$, for each model. Note that even at $\tau = 0$ for the tested model, some variance across the 20 repetitions of a given game can remain, since we vary only the tested model's temperature: the verifier and extractor (Qwen) continue to run at their fixed setting of $\tau = 0.3$ (Methods Subsection~\ref{subsec:modelparam}), and the label-to-strategy assignment (which letter, A or B, denotes cooperation) is independently randomized across repetitions (Methods Subsection~\ref{subsec:method_adapt}). Both of these factors can introduce variance into the recorded cooperation outcome independently of the tested model's own temperature.  Figure~\ref{fig:tempvariance} shows that increasing temperature increases the variance of cooperation rates within each game's 20 repetitions, as expected. Figure~\ref{fig:tempperformance} shows that this increased variance does not come at the cost of accuracy: point estimates of model-human and model-Nash similarity remain broadly stable across temperatures from $0.1$ to $1$, while all three models perform noticeably worse at $\tau=0$. We use $\tau = 0.8$ for the main results reported throughout the paper.

\vspace{0.8em}

\vspace{0.2em}

\subsection{Prior experimental studies with human participants using game-theoretic payoff structures} \label{app:prior_studies}

Table~\ref{tab:prior_studies} provides a selection of prior experimental studies using game-theoretic payoff structures that may overlap with the parameter space explored in our simulations. The studies span several decades, multiple game types, a wide range of payoff scales, and both one-shot and repeated-play designs. Cooperation rates vary substantially even across structurally similar games, illustrating that any memorization-based account of LLM behavior would require
non-trivial abstraction across heterogeneous sources rather than direct retrieval.

\input{prior_studies_table.tex}

\subsection{Full-text of prompts}
\label{app:subsec:prompts}

This subsection provides the complete, literal text of every prompt referenced abstractly in Methods Sections~\ref{subsec:modelparam}--\ref{subsec:method_extract}. The Shared Components below correspond to the system and user prompt roles described in Subsection~\ref{subsec:modelparam}: a fixed system message, and the user prompt content, comprising the game instructions, the payoff table, and (at later stages) the multi-step reasoning scaffold appended to it. The four Stage subsections that follow then show exactly how these shared components, together with the Qwen extraction prompt and the logical verifier prompt, are assembled into the complete message sequence for each of the four extraction methods introduced in Subsection~\ref{subsec:method_extract}: Simple, Double (here labeled \texttt{extract}), Multi-step, and Logical Verifier (here labeled \texttt{final}).

\lstset{breakindent=0pt}
% 

% ============================================================
\fakesec{Shared Components}
% ============================================================

\fakesub{Main model system message (all stages)}

\begin{lstlisting}
You are a helpful assistant.
\end{lstlisting}

\fakesub{Instructions (all stages, User turn 1)}

\begin{lstlisting}
This one-shot game is made of only one round with an anonymous player. You will play only once.

To play you must choose one of two options: A and B, the other player will also choose between A and B. Both players are subjected to the same rules.

You won't know the other player's decision until the end of the round so you won't be able to change your choice after knowing the other player's decision.
The other player won't know your decision until the end of the round so they won't be able to change their choice after knowing yours.

You will be playing simultaneously with the other player.

Both players will receive a monetary prize. The monetary prize for the player will be the amount of points the player has earned multiplied by 10.

For example, if player 1 earns 9 points and player 2 earns 5 points.
Player 1 will receive 9*10 = 90 euros and player 2 will receive 10*5 = 50 euros.

The monetary prize of each player only depends on their number of points.
The difference in points of both players has no effect on the prize.

Both players will receive the prize, regardless of who earned more or less money.
The outcome is solely determined by the number of points earned by each player.
\end{lstlisting}

\fakesub{Payoff table --- script-0 variant (A = cooperate)}

\begin{lstlisting}
If you choose A and the other player chooses A. You earn 10 points, the other player earns 10 points.
If you choose A and the other player chooses B. You earn {s} points, the other player earns {t} points.
If you choose B and the other player chooses A. You earn {t} points, the other player earns {s} points.
If you choose B and the other player chooses B. You earn 5 points, the other player earns 5 points.
\end{lstlisting}

\fakesub{Payoff table --- script-1 variant (B = cooperate)}

\begin{lstlisting}
If you choose B and the other player chooses B. You earn 10 points, the other player earns 10 points.
If you choose B and the other player chooses A. You earn {s} points, the other player earns {t} points.
If you choose A and the other player chooses B. You earn {t} points, the other player earns {s} points.
If you choose A and the other player chooses A. You earn 5 points, the other player earns 5 points.
\end{lstlisting}

\fakesub{Reasoning steps (multi and final stages, appended to payoff table)}

\begin{lstlisting}
1. There are four possible outcomes in this game, depending on your choice (A or B) and the other player's choice (A or B).
2. Group the outcomes based on your decision:
   - If you choose A: (A,A) and (A,B)
   - If you choose B: (B,A) and (B,B)
3. Analyze and compare the outcomes in each group step by step.
4. Make sure your analysis is consistent with the rules of the game and the possible outcomes.
5. Based on your analysis, choose either A or B.
\end{lstlisting}

% ============================================================
\fakesec{Stage: simple}
% ============================================================

One call to the main model. No extractor. Uses the \texttt{\_q} payoff variants
(which include the closing question) plus the constrained suffix.

\fakesub{System message}

\begin{lstlisting}
[main model system message -- see above]
\end{lstlisting}

\fakesub{User turn 1}

\begin{lstlisting}
[instructions -- see above]
\end{lstlisting}

\fakesub{User turn 2 --- script-0-q variant}

\begin{lstlisting}
[payoff table script-0 -- see above]

What do you choose? A or B?

Choose between A and B. Answer only with A or B depending on your choice.
\end{lstlisting}

\fakesub{User turn 2 --- script-1-q variant}

\begin{lstlisting}
[payoff table script-1 -- see above]

What do you choose? A or B?

Choose between A and B. Answer only with A or B depending on your choice.
\end{lstlisting}

% ============================================================
\fakesec{Stage: extract}
% ============================================================

Two calls: main model, then Qwen extractor.

\fakesub{Main model --- System}

\begin{lstlisting}
[main model system message -- see above]
\end{lstlisting}

\fakesub{Main model --- User turn 1}

\begin{lstlisting}
[instructions -- see above]
\end{lstlisting}

\fakesub{Main model --- User turn 2 (script-0)}

\begin{lstlisting}
[payoff table script-0 -- see above]

Make a choice and explain why
\end{lstlisting}

\fakesub{Main model --- User turn 2 (script-1)}

\begin{lstlisting}
[payoff table script-1 -- see above]

Make a choice and explain why
\end{lstlisting}

\fakesub{Qwen extractor --- System}

\begin{lstlisting}
You are a helpful assistant.
\end{lstlisting}

\fakesub{Qwen extractor --- User turn 1}

\begin{lstlisting}
The player who was asked to choose between A and B answered: {main_model_output}
\end{lstlisting}

\fakesub{Qwen extractor --- User turn 2}

\begin{lstlisting}
What did the person who wrote the message choose? Answer shortly.
\end{lstlisting}

% ============================================================
\fakesec{Stage: multi}
% ============================================================

Same as extract but reasoning steps are appended to the payoff table
instead of the free-form suffix.

\fakesub{Main model --- System}

\begin{lstlisting}
[main model system message -- see above]
\end{lstlisting}

\fakesub{Main model --- User turn 1}

\begin{lstlisting}
[instructions -- see above]
\end{lstlisting}

\fakesub{Main model --- User turn 2 (script-0)}

\begin{lstlisting}
[payoff table script-0 -- see above]

[reasoning steps -- see above]
\end{lstlisting}

\fakesub{Main model --- User turn 2 (script-1)}

\begin{lstlisting}
[payoff table script-1 -- see above]

[reasoning steps -- see above]
\end{lstlisting}

\fakesub{Qwen extractor}

\begin{lstlisting}
[same as extract stage -- see above]
\end{lstlisting}

% ============================================================
\fakesec{Stage: final}
% ============================================================

Three calls per iteration: main model (same as multi), then Qwen verifier,
then Qwen extractor (only for games that passed the verifier).

\fakesub{Main model}

\begin{lstlisting}
[same prompts as multi stage -- see above]
\end{lstlisting}

\fakesub{Qwen verifier --- System}

\begin{lstlisting}
You are a logical verifier. Your purpose is to look for inconsistencies and errors in a given text.
\end{lstlisting}

\fakesub{Qwen verifier --- User turn}

The verifier message is assembled as:
\texttt{\_VERIFIER\_CRITERIA} + \texttt{instructions\_script\_short} + 5 sampled examples + footer.

\subsubsection*{Criteria block (\texttt{\_VERIFIER\_CRITERIA})}

\begin{lstlisting}
You have to read the answer of Player 1 and output good or bad according to the following criteria:

Criteria for good:
- ALL arithmetic/mathematical comparisons and claims are CORRECT.
- ALL descriptions of the possible outcomes are CORRECT.
- ONLY TWO possible outputs per choice, for A: (A,A) and (A,B), for B: (B,A) and (B,B).
- The final choice is CONSISTENT with the previous reasoning.
- ALL statements make sense.
- GOOD UNDERSTANDING of the rules.

Criteria for bad:
- ONE OR MORE arithmetic/mathematical comparisons or claims are WRONG.
- ONE OR MORE descriptions of the possible outcomes are WRONG.
- Does not understand that there are ONLY TWO possible outcomes per choice.
- Final choice is NOT CONSISTENT with the previous reasoning.
- ONE OR MORE statements do not make sense.
- BAD UNDERSTANDING of the rules.

Rules of the game:
\end{lstlisting}

\subsubsection*{Instructions block (\texttt{instructions\_script\_short})}

\begin{lstlisting}
[instructions -- see above]
\end{lstlisting}

\subsubsection*{Example block (repeated for each of 5 sampled examples)}

\begin{lstlisting}
Example {i}, {good|bad} reasoning:

Rules of Example {i}:

{rules text}

Answer of Example {i}:

{answer text}

Output of Example {i}: {good|bad}
\end{lstlisting}

\subsubsection*{Footer block}

\begin{lstlisting}
Rules of the game to analyze:
{points}

Answer of Player 1, to analyze:
{answer1}

Important:
- Output ONLY one word: good or bad
- Do not add punctuation, extra spaces, or explanations.

\end{lstlisting}

\fakesub{Verifier example pool --- Example 1 (bad, S=7 T=6, chose B when A strictly dominates)}

\begin{lstlisting}
If you choose A and the other player chooses A. You earn 10 points, the other player earns 10 points.
If you choose A and the other player chooses B. You earn 7 points, the other player earns 6 points.
If you choose B and the other player chooses A. You earn 6 points, the other player earns 7 points.
If you choose B and the other player chooses B. You earn 5 points, the other player earns 5 points.
\end{lstlisting}

\begin{lstlisting}
Based on the rules and possible outcomes, here's the analysis:

- If you choose A, you can earn either 10 points (if the other player chooses A) or 7 points (if the other player chooses B).
- If you choose B, you can earn either 5 points (if the other player chooses B) or 6 points (if the other player chooses A).

However, the rules state that the difference in points between you and the other player has no effect on the prize. Therefore, it comes down to the following:

- If you choose A, the highest possible prize is 10 points, but the lowest possible prize is 7 points.
- If you choose B, the highest possible prize is 5 points, but the lowest possible prize is 6 points.

Since the goal is to maximize the possible prize, it would seem that choosing A gives you the potential for a higher prize (100 euros) compared to choosing B (60 euros as the highest potential prize). However, it's important to keep in mind that the other player is making the same analysis as well, so it might be wiser to choose B to reduce the chances of both players choosing A and thus reducing the potential prize for both.

As this is a one-shot game, it's difficult to say for certain which choice will lead to a higher average prize for both players. It could be a game of bluff and strategy rather than a purely mathematical decision.

In this analysis, I would lean towards choosing B, to potentially lower the chances of both players receiving a lower prize, but it's ultimately up to your personal decision.
\end{lstlisting}

\noindent Output: \texttt{bad}

\fakesub{Verifier example pool --- Example 2 (good, S=10 T=6, chose A cooperate)}

\begin{lstlisting}
If you choose A and the other player chooses A. You earn 10 points, the other player earns 10 points.
If you choose A and the other player chooses B. You earn 10 points, the other player earns 6 points.
If you choose B and the other player chooses A. You earn 6 points, the other player earns 10 points.
If you choose B and the other player chooses B. You earn 5 points, the other player earns 5 points.
\end{lstlisting}

\begin{lstlisting}
To make a decision, let's analyze the outcomes based on the rules and possible outcomes:

1. If you choose A:
   - (A,A): Both players earn 10 points each.
   - (A,B): You earn 10 points, and the other player earns 6 points.

2. If you choose B:
   - (B,A): You earn 6 points, and the other player earns 10 points.
   - (B,B): Both players earn 5 points each.

3. Since the monetary prize for each player is the number of points multiplied by 10, it is clear that earning more points results in a higher prize.

4. Comparing the outcomes in each group:
   - If you choose A: (A,A) gives you the highest reward, and (A,B) gives you the same reward but a lower-one for the other player.
   - If you choose B: (B,B) gives you the lowest reward, and (B,A) gives a slightly higher reward, but still lower than all possible rewards when choosing A.

5. Based on this information, choosing A would result in a higher potential reward, making it the better choice.
Even though one might worry about the other player choice, it doesn't make sense since the only possible outcome of choosing A is 10 points, which is the maximum.
The other player choice isn't going to affect your outcome at all, there is no risk.
\end{lstlisting}

\noindent Output: \texttt{good}

\fakesub{Verifier example pool --- Example 3 (bad, S=9 T=15, fabricated probabilities)}

\begin{lstlisting}
If you choose A and the other player chooses A. You earn 10 points, the other player earns 10 points.
If you choose A and the other player chooses B. You earn 9 points, the other player earns 15 points.
If you choose B and the other player chooses A. You earn 15 points, the other player earns 9 points.
If you choose B and the other player chooses B. You earn 5 points, the other player earns 5 points.
\end{lstlisting}

\begin{lstlisting}
Let's analyze the outcomes:
- If you choose A, the probability of the other player also choosing A is 50%.
In this case, you would earn 10 points, and the other player would also earn 10 points (outcome A,A).
The other possibility is that the other player chooses B, in which case you would earn 9 points, and they would earn 15 points (outcome A,B).
- If you choose B, the probability of the other player also choosing B is 25%.
In this case, you would earn 5 points, and the other player would also earn 5 points (outcome B,B).
The other possibility is that the other player chooses A, in which case you would earn 15 points, and they would earn 9 points (outcome B,A).

Now, let's compare the outcomes:
- If you choose A, you have two possible outcomes: one with a higher reward (10 points) and one with a lower reward (9 points).
- If you choose B, you have two possible outcomes: one with a lower reward (5 points) and one with a higher reward (15 points).

Since you don't know the other player's choice, it's impossible to determine the exact outcome.
However, it's interesting to note that if the other player chooses A, choosing A would result in a lower reward (9 points) compared to choosing B (15 points).
On the other hand, if the other player chooses B, choosing A would result in a higher reward (10 points) compared to choosing B (5 points).
Given this analysis, it seems that choosing B gives you the potential for a higher reward, but it's also possible that the other player will choose A in which case you would have a lower reward.
It's truly a gamble. In this game, there is no optimal strategy as the game relies on the other player's choice, which remains unknown.
Therefore, it's hard to make a definitive choice between A and B as there is no clear advantage in either option.
Based on the analysis and the lack of a perfect answer, I'll leave the choice up to you.
You can choose either A or B as there's no guaranteed outcome. Good luck!
\end{lstlisting}

\noindent Output: \texttt{bad}

\fakesub{Verifier example pool --- Example 4 (good, S=8 T=5, chose B cooperate in script-1)}

\begin{lstlisting}
If you choose B and the other player chooses B. You earn 10 points, the other player earns 10 points.
If you choose B and the other player chooses A. You earn 8 points, the other player earns 5 points.
If you choose A and the other player chooses B. You earn 5 points, the other player earns 8 points.
If you choose A and the other player chooses A. You earn 5 points, the other player earns 5 points.
\end{lstlisting}

\begin{lstlisting}
Based on the given information, I will analyze the outcomes for each choice:

If you choose A:
- If the other player chooses A, you both will receive 5 points each.
- If the other player chooses B, you will receive 5 points, while the other player receives 8 points.

If you choose B:
- If the other player chooses A, you will receive 8 points, while the other player receives 5 points.
- If the other player chooses B, you both will receive 10 points each.

If you choose B, the maximum amount of points you can get is 10 points and the minimum, 8 points.
If you choose A the maximum amount of points you can get is 5 points and the minimum, 5 points.
The maximum points is greater when you choose B, leading to a higher prize.
The minimum points is greater when you choose B, leading to a lower risk of getting little points.
Then, choice B is optimal since it is the option with the highest reward and the lower risk.
\end{lstlisting}

\noindent Output: \texttt{good}

\fakesub{Verifier example pool --- Example 5 (bad, S=10 T=7, incomplete reasoning in script-1)}

\begin{lstlisting}
If you choose B and the other player chooses B. You earn 10 points, the other player earns 10 points.
If you choose B and the other player chooses A. You earn 10 points, the other player earns 7 points.
If you choose A and the other player chooses B. You earn 6 points, the other player earns 10 points.
If you choose A and the other player chooses A. You earn 5 points, the other player earns 5 points.
\end{lstlisting}

\begin{lstlisting}
Given the rules, choosing A would bring a maximum of 5 points, and choosing B would bring a maximum of 10 points if the other player also chooses B.
Since we don't know the other player's choice, it's impossible to guarantee a higher score by choosing A or B.
However, if we're forced to make a choice, choosing B seems like the better option since it has a higher potential reward if the other player also chooses B.
But, ultimately, this game is a gamble, and both choices have an element of risk.
\end{lstlisting}

\noindent Output: \texttt{bad}

\fakesub{Verifier example pool --- Example 6 (good, S=1 T=15, chose B defect)}

\begin{lstlisting}
If you choose A and the other player chooses A. You earn 10 points, the other player earns 10 points.
If you choose A and the other player chooses B. You earn 1 points, the other player earns 15 points.
If you choose B and the other player chooses A. You earn 15 points, the other player earns 1 points.
If you choose B and the other player chooses B. You earn 5 points, the other player earns 5 points.
\end{lstlisting}

\begin{lstlisting}
Choosing A can lead to the worst result (earning only 1 point) and choosing B can lead to the best result (earning 15 points).
Then I choose B to earn at least 5 points and also have the potential for the highest reward.
\end{lstlisting}

\noindent Output: \texttt{good}

\fakesub{Verifier example pool --- Example 7 (good, S=0 T=12, chose B defect)}

\begin{lstlisting}
If you choose A and the other player chooses A. You earn 10 points, the other player earns 10 points.
If you choose A and the other player chooses B. You earn 0 points, the other player earns 12 points.
If you choose B and the other player chooses A. You earn 12 points, the other player earns 0 points.
If you choose B and the other player chooses B. You earn 5 points, the other player earns 5 points.
\end{lstlisting}

\begin{lstlisting}
Based on the rules, let me analyze each possible choice:

1. If I choose A:
   - (A,A): I earn 10 points.
   - (A,B): I earn 0 points.

2. If I choose B:
   - (B,A): I earn 12 points.
   - (B,B): I earn 5 points.

Comparing the outcomes: when the other player chooses A, B yields 12 points versus 10 for A. When the other player chooses B, B yields 5 points versus 0 for A. In both scenarios B gives a higher payoff than A. B is therefore the dominant strategy regardless of what the other player does, and I choose B.
\end{lstlisting}

\noindent Output: \texttt{good}

\fakesub{Verifier example pool --- Example 8 (bad, S=2 T=12, chose A cooperate when B strictly dominates)}

\begin{lstlisting}
If you choose A and the other player chooses A. You earn 10 points, the other player earns 10 points.
If you choose A and the other player chooses B. You earn 2 points, the other player earns 12 points.
If you choose B and the other player chooses A. You earn 12 points, the other player earns 2 points.
If you choose B and the other player chooses B. You earn 5 points, the other player earns 5 points.
\end{lstlisting}

\begin{lstlisting}
Both players want to maximize their points. If both choose A, we both earn 10 points, which is fair and gives the best combined outcome of 20 points total. Choosing B is risky: if both players choose B we only earn 5 each, which is worse for everyone. Since we are rational players and cooperation leads to the best mutual outcome, I choose A.
\end{lstlisting}

\noindent Output: \texttt{bad}

\fakesub{Verifier example pool --- Example 9 (good, S=4 T=13, chose A defect in script-1)}

\begin{lstlisting}
If you choose B and the other player chooses B. You earn 10 points, the other player earns 10 points.
If you choose B and the other player chooses A. You earn 4 points, the other player earns 13 points.
If you choose A and the other player chooses B. You earn 13 points, the other player earns 4 points.
If you choose A and the other player chooses A. You earn 5 points, the other player earns 5 points.
\end{lstlisting}

\begin{lstlisting}
Let me compare the two possible choices:

1. If I choose A:
   - (A,B): I earn 13 points.
   - (A,A): I earn 5 points.

2. If I choose B:
   - (B,B): I earn 10 points.
   - (B,A): I earn 4 points.

Comparing outcome by outcome: if the other player chooses B, A gives me 13 while B gives me 10. If the other player chooses A, A gives me 5 while B gives me 4. In both cases A yields a strictly higher reward. A dominates B regardless of the other player's choice, so I choose A.
\end{lstlisting}

\noindent Output: \texttt{good}

\fakesub{Verifier example pool --- Example 10 (bad, S=3 T=11, chose B cooperate when A strictly dominates)}

\begin{lstlisting}
If you choose B and the other player chooses B. You earn 10 points, the other player earns 10 points.
If you choose B and the other player chooses A. You earn 3 points, the other player earns 11 points.
If you choose A and the other player chooses B. You earn 11 points, the other player earns 3 points.
If you choose A and the other player chooses A. You earn 5 points, the other player earns 5 points.
\end{lstlisting}

\begin{lstlisting}
This game presents a trade-off between individual gain and mutual benefit. If both players choose B, we each earn 10 points - a fair and cooperative outcome. If I choose A and the other player chooses B, I gain 11 but the other player earns only 3, which is selfish. Since we are both rational players and the rules are symmetric, the other player is likely reasoning the same way. We should both prefer the cooperative equilibrium where we each earn 10 points. I choose B.
\end{lstlisting}

\noindent Output: \texttt{bad}

\fakesub{Qwen extractor (final stage)}

\begin{lstlisting}
[same as extract/multi stages -- see above; only called for games that passed the verifier]
\end{lstlisting}

\subsection{\label{app:subsec:othermodels}Prompting stages for Qwen and Mistral}

\begin{figure}[htpb!]
    \centering
    \includegraphics[width=0.9\linewidth]{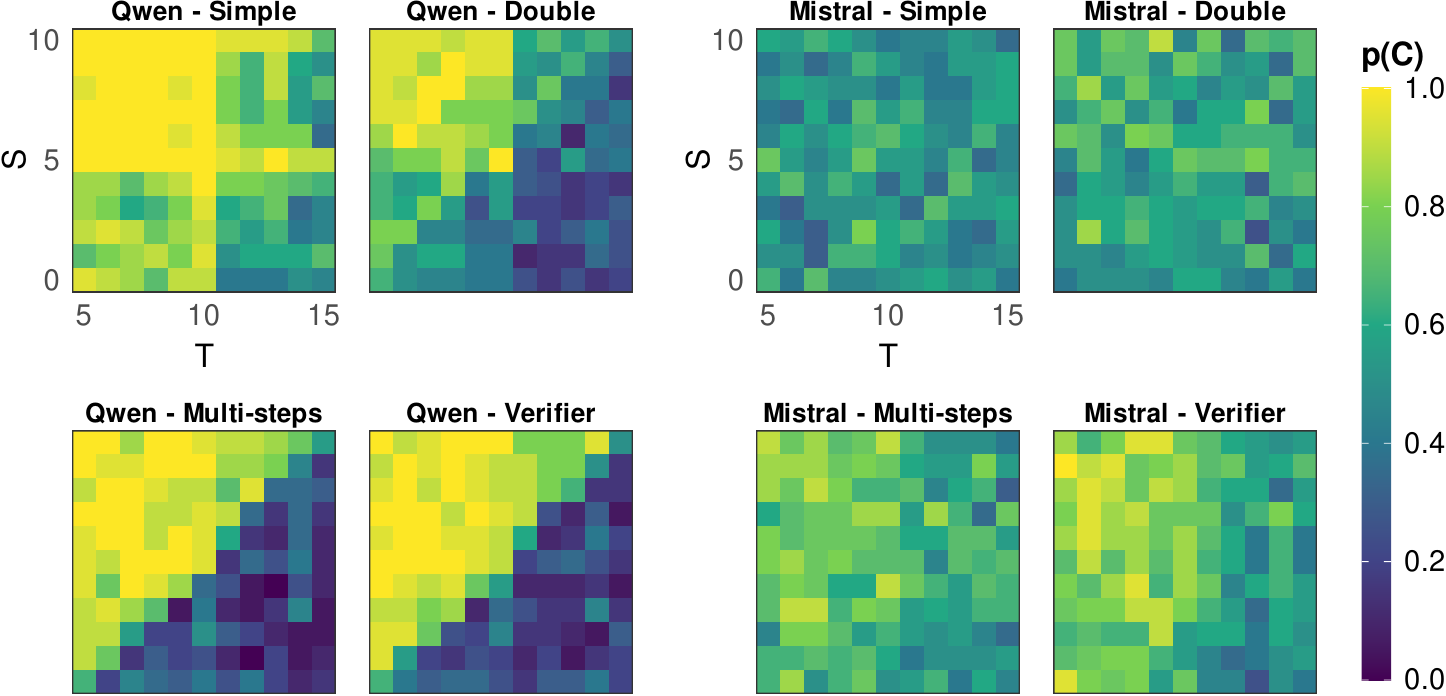}
    \caption{\textbf{Progressive changes in Mistral's and Qwen's cooperation matrices with increasingly sophisticated prompting methods.} Each panel shows results from Simple Extraction, Double Extraction, Multi-step Extraction, and Logical Verifier Extraction (from top to bottom for each model). Color scales indicate average cooperation ranging from 0 (purple: no cooperation) to 1 (yellow: full cooperation). Qwen exhibits stable patterns earlier in the extraction process, while Mistral requires more complex methods to reveal its cooperation behavior.}
    \label{app:fig:methodsmistralqwen}
\end{figure}

While the main analysis focuses on results obtained with Llama, we also conducted the same simulations of the prompting stages we use with Qwen and Mistral. As shown in Figure~\ref{app:fig:methodsmistralqwen}, progressively more sophisticated decision extraction methods reveal clearer cooperation patterns in both models, mirroring the improvements observed with Llama. Qwen exhibits roughly a cooperation matrix with the four distinct game regions even with Simple Extraction, and from Double Extraction on we obtain a stable pattern. In contrast, Mistral is approaching a non-random cooperation matrix only from the Multi-steps extraction stage, demonstrating greater sensitivity to the extraction methodology.

\subsection{Payoff Saliency of Qwen and Mistral}
\label{app:subsec:attention_othermodels}

\begin{figure}[H]
    \centering
    \includegraphics[width=0.75\linewidth]{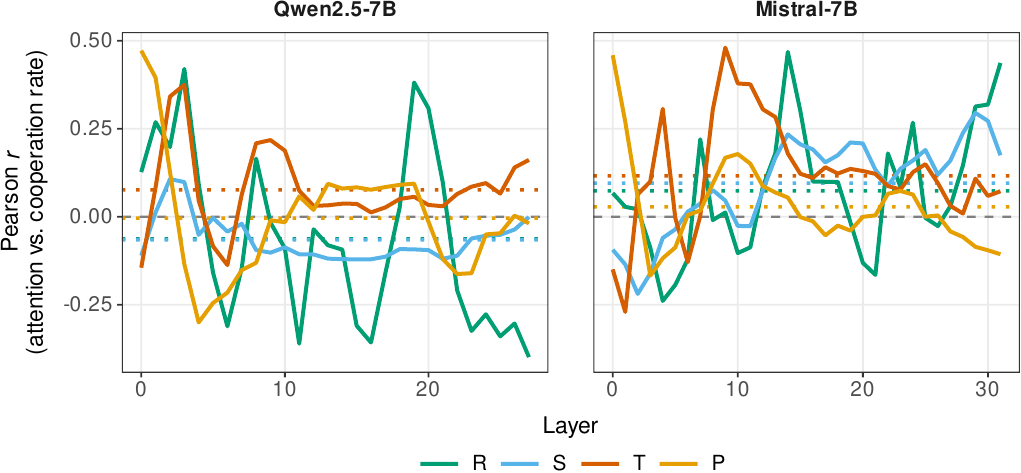}
    \caption{\textbf{Payoff salience for Qwen and Mistral: attention to payoff values vs. cooperation rate.} Equivalent analysis to Figure~\ref{fig:attention_analysis}, shown here for Qwen2.5-7B (left) and Mistral-7B (right). Neither model exhibits Llama's structured, layer-dependent organization of payoff salience: Qwen shows no sustained signal for any payoff across network depth, and Mistral shows only a weak, noisy echo of the R-positive correlation seen in Llama.}
    \label{fig:attention_othermodels}
\end{figure}

While Llama exhibits a structured, layer-dependent payoff saliency profile (Figure~\ref{fig:attention_analysis}), neither Qwen nor Mistral show a comparable pattern. Qwen's attention-cooperation correlations (Figure~\ref{fig:attention_othermodels}, left) oscillate around zero for all four payoffs across network depth, without the sustained, anti-correlated R/T trajectory seen in Llama; no payoff shows a consistent directional signal across layers. Mistral (Figure~\ref{fig:attention_othermodels}, right) shows a noisier profile in which T and R both display brief positive peaks at different depths, but neither sustains the kind of structured, opposing trajectory that characterizes Llama, and S and P show no consistent signal either. This absence of structure in both models is consistent with their comparatively weaker behavioral alignment with human cooperation patterns (Table~\ref{combined_table}), and supports the interpretation that Llama's closer behavioral fit to humans is accompanied by a distinctive internal organization of payoff information that is not shared by the other two models.

\subsection{Extending games with Qwen and Mistral}
\label{app:subsec:extension_othermodels}

\begin{figure}
    \centering
    \includegraphics[width=0.96\linewidth]{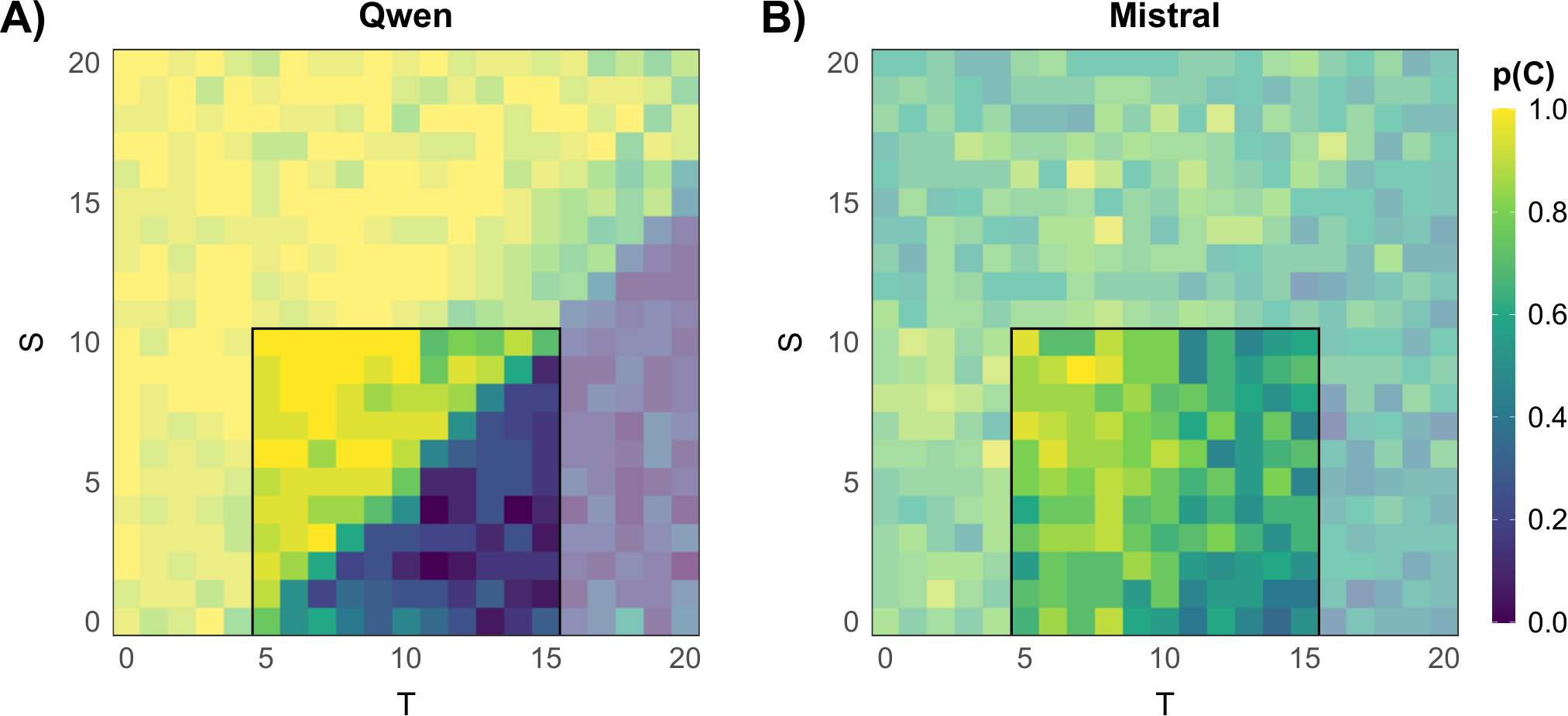}
    \caption{\textbf{Cooperation matrices of Qwen and Mistral, extended through simulations.} Each tile represents a game defined by the parameter combination of $(S,T)$, with average cooperation ranging from $0$ (purple: no cooperation) to $1$ (yellow: full cooperation). Panel A: Cooperation matrix of Qwen. Panel B: Cooperation matrix of Mistral. In both panels, the non-transparent region delineated by a black border shows simulations with parameter combinations that were already part of the original human experiments; the transparent region shows combinations of $S$ and $T$ that were not played by the human participants. Compare to Llama's extended cooperation matrix and the Nash equilibrium predictions in Figure~\ref{fig:novel_games}.}
    \label{fig:extendedgames_othermodels}
\end{figure}

Using the same extended experimental setup as for Llama (Subsection~\ref{subsec:novelgames}), we simulated Qwen and Mistral across the full $441$-game grid. Panel A of Figure~\ref{fig:extendedgames_othermodels} shows Qwen's extended cooperation matrix. Cooperation remains high in the $S \geq T$ region and falls substantially when $T$ exceeds $R$, preserving the same general structure observed within the original 121-game grid and consistent with Qwen's close alignment with Nash equilibrium predictions (Table~\ref{combined_table}). Unlike Llama, whose cooperation-defection boundary tracks closely along the $S=T$ diagonal, Qwen's boundary in the extended grid sits systematically below this diagonal, continuing to cooperate well into the region where $S<T$. This means Qwen's decisions are not simply governed by the comparison between $S$ and $T$, as in the Envious phenotype that characterizes Llama, but follow a different rule, one more in line with Qwen's closer overall alignment with Nash equilibrium predictions (Table~\ref{combined_table}) than with this simpler heuristic. Panel B shows Mistral's extended cooperation matrix, which is markedly less differentiated across the grid than either Llama's or Qwen's. Notably, mean cooperation in the $S \geq T$ region is similar to the $S<T$, $T \leq R$ region, in contrast to the clear $S \geq T$-dominant pattern seen in Llama and Qwen. We find a substantially higher rate of unverified responses for Mistral (described in Subsection~\ref{subsec:problemgames}), which suggests that its extended cooperation matrix is shaped less by the strategic structure of the game than is the case for the other two models.

\subsection{Problematic Games}\label{subsec:problemgames}

\begin{figure}
    \centering
    \includegraphics[width=0.88\linewidth]{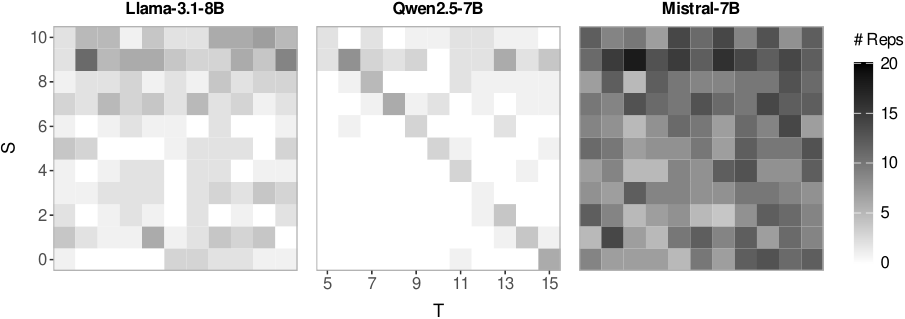}
    \caption{\textbf{Logical verifier bypass across game space.} Number of repetitions (out of 20) in which the logical verifier was bypassed per game. Bypass is triggered after three consecutive stuck iterations for a given game; that game's response is then accepted without quality verification.}
    \label{fig:problematic_games}
\end{figure}

When the logical verifier fails to accept any response for a given game across three consecutive iterations, it deactivates for that game's remaining responses, which proceed directly to answer extraction. Cooperation rates from these less-filtered games may reflect logical inconsistencies or mathematical errors rather than genuine strategic reasoning. Figure~\ref{fig:problematic_games} shows the fraction of responses for each game that bypassed the logical verifier. The pattern confirms that games with higher values of $S$ and $T$ are more problematic, producing fewer logically correct answers and thus yielding cooperation estimates with potentially higher error (correlation between bypass rate and $S$: $r=0.45$ for Llama, $r=0.32$ for Qwen, $r=0.40$ for Mistral; with $T$: $r=0.08$, $r=0.11$, $r=0.24$ respectively). The extent of this effect differs substantially across models. For Llama, the mean bypass rate across all 121 games is $11.2\%$ (median $10\%$), with the single most affected game reaching $55\%$. Qwen is least affected, with a mean bypass rate of $4.3\%$ (median $0\%$) and a maximum of $40\%$. Mistral is affected far more substantially, with a mean bypass rate of $49.3\%$ across all games (median $50\%$) and several individual games exceeding $75\%$, including one game in which $90\%$ of repetitions bypassed verification. Cooperation estimates for Mistral should therefore be interpreted with this caveat: roughly half of its recorded responses across the full game space did not pass logical verification, a substantially larger share than for Llama or Qwen.

\subsection{Details on games and behavioral experiments}
\label{app:subsec:detailgames}

\begin{itemize}
    \item \textbf{Harmony game.} A two-player game where cooperation is the best choice for each player regardless of what the other player does. Both players receive higher payoffs by cooperating than by defecting, and even if exploited by a defector, a cooperator still fares better than in mutual defection. This game represents situations where individual incentives perfectly align with collective welfare, measuring unconditional cooperation in scenarios with no strategic tension.
    \item \textbf{Stag Hunt game.} A coordination game where players achieve the highest joint payoff through mutual cooperation, but face significant risk if they cooperate while their partner defects. Mutual defection provides a safe but suboptimal outcome. Players must balance the appeal of higher collective rewards against the security of risk-averse choices, creating two stable equilibria. This game captures situations requiring trust and coordinated commitment, measuring players' willingness to pursue collectively optimal but individually risky outcomes \cite{skyrmsStagHunt2001}.
    \item \textbf{Snowdrift game.} Also known as Hawk-Dove or Chicken game, where players benefit most when their opponent cooperates while they defect, but mutual defection produces the worst outcome for both. Unlike the Prisoner's Dilemma, being exploited is preferable to mutual defection, creating incentives for anti-coordination. This game models scenarios where avoiding the worst outcome encourages at least one player to cooperate, measuring tolerance for asymmetric contributions and strategic adaptation \cite{rapoportGameChicken1966}.
    \item \textbf{Prisoner's dilemma.} This game consists of two players who can either cooperate or defect. If one player defects while the other cooperates, they receive the best outcome, and the other, the worst. If both players cooperate, they receive a higher joint payoff. Humans \cite{rapoport_1965} show a tendency toward cooperation and are influenced by past outcomes. This game measures collective well-being versus personal interest.
    \item \textbf{Dictator game.} One player is given an amount of money and can decide how much to keep and how much to give to the other player. Humans tend to keep most of the money\cite{kahneman_1986}. This game is used to measure altruism.
    \item \textbf{Ultimatum game.} This game is similar to the Dictator game, but the recipient can reject the offer, resulting in neither player receiving anything. In humans,\cite{guth_1982} the allocations are more balanced than in the Dictator game due to fear of punishment. This game is used to measure fairness, altruism, and responses to unfairness.
    \item \textbf{Trust game.} One player, the investor, receives a quantity of money and decides how much to keep and how much to give to the other player, the banker. The banker multiplies the received money by a factor and decides how much to return to the investor. Humans \cite{berg_1995} tend to show reciprocity. This game measures altruism and fairness. 
    \item \textbf{Wisdom of crowds.} An observed phenomenon whereby averaging the responses to a question yields more accurate results than individual responses \cite{hong_2004}. This phenomenon measures collective intelligence.
    \item \textbf{Milgram obedience experiment.} This experiment involves an authority figure, a participant who must obey, and a subject who experiences pain (simulated electric shocks) whenever the authority figure dictates, administered by the obedient participant. It measures obedience to authority. In humans \cite{milgram_1963}, obedience remains high until the subject receiving pain begins to complain.
\end{itemize}

\end{document}

%% file: metrics_table.tex
\begin{tabular}{lcccc}
\toprule
& \multicolumn{2}{c}{\textbf{Human}} & \multicolumn{2}{c}{\textbf{Nash}} \\
\cmidrule(lr){2-3} \cmidrule(lr){4-5}
& \textbf{MSD} & $\boldsymbol{r}$ & \textbf{MSD} & $\boldsymbol{r}$ \\
\midrule
\textbf{Llama} &
  \shortstack{\rule{0pt}{12pt}\textbf{0.020} \\[3pt] {\small \textcolor{gray}{[0.015, 0.025]}}}
& \shortstack{\rule{0pt}{12pt}\textbf{0.84} \\[3pt] {\small \textcolor{gray}{[0.78, 0.89]}}}
& \shortstack{\rule{0pt}{12pt}0.104 \\[3pt] {\small \textcolor{gray}{[0.083, 0.125]}}}
& \shortstack{\rule{0pt}{12pt}0.69 \\[3pt] {\small \textcolor{gray}{[0.59, 0.77]}}} \\[4pt]
\textbf{Qwen} &
  \shortstack{\rule{0pt}{12pt}0.068 \\[3pt] {\small \textcolor{gray}{[0.053, 0.083]}}}
& \shortstack{\rule{0pt}{12pt}0.77 \\[3pt] {\small \textcolor{gray}{[0.68, 0.83]}}}
& \shortstack{\rule{0pt}{12pt}\textbf{0.033} \\[3pt] {\small \textcolor{gray}{[0.025, 0.041]}}}
& \shortstack{\rule{0pt}{12pt}\textbf{0.93} \\[3pt] {\small \textcolor{gray}{[0.91, 0.95]}}} \\[4pt]
\textbf{Mistral} &
  \shortstack{\rule{0pt}{12pt}0.063 \\[3pt] {\small \textcolor{gray}{[0.052, 0.074]}}}
& \shortstack{\rule{0pt}{12pt}0.67 \\[3pt] {\small \textcolor{gray}{[0.56, 0.76]}}}
& \shortstack{\rule{0pt}{12pt}0.168 \\[3pt] {\small \textcolor{gray}{[0.134, 0.202]}}}
& \shortstack{\rule{0pt}{12pt}0.60 \\[3pt] {\small \textcolor{gray}{[0.47, 0.70]}}} \\[4pt]
\midrule
\textbf{Nash} &
  \shortstack{\rule{0pt}{12pt}0.095 \\[3pt] {\small \textcolor{gray}{[0.078, 0.111]}}}
& \shortstack{\rule{0pt}{12pt}0.78 \\[3pt] {\small \textcolor{gray}{[0.70, 0.84]}}} \\[4pt]
\bottomrule
\end{tabular}

%% file: prior_studies_table.tex
% Table: A selection of prior experimental studies
% Place in appendix, referenced from memorization discussion
%
% Requires: booktabs, multirow, array in preamble

\begin{table}[p]
\renewcommand{\arraystretch}{1.1}
\centering
\footnotesize
\caption{\textbf{A selection of prior experimental studies using game-theoretic payoff structures.} Studies span several decades, a wide range of payoff scales, multiple game types,
and both one-shot and repeated-play designs. Cooperation rates vary substantially even for structurally similar games. This heterogeneity illustrates why simple memorization of prior experimental results would not straightforwardly explain LLM behavior in our setup (R = 10, P = 5). A dash (--) indicates that aggregate cooperation rates were not available from the source.}
\label{tab:prior_studies}
\setlength{\tabcolsep}{5pt}
\begin{tabular}{llcrrrrr}
\toprule
\textbf{Study} & \textbf{Type} & \textbf{Context} & \textbf{R} & \textbf{S} & \textbf{T} & \textbf{P} & \textbf{Avg.\ $p(C)$} \\
\midrule
\multicolumn{8}{l}{\textit{Prisoner's Dilemma --- one-shot}} \\[2pt]

Frank \& Gilovich (1993)\cite{frankDoesStudyingEconomics1993} & PD & Lab & 2 & 0 & 3 & 1 & 0.40--0.61 \\[4pt]

Andreoni \& Miller (1993)\cite{andreoniRationalCooperationFinitely1993} & PD & Lab & 7 & 0 & 12 & 4 & 0.18 \\[4pt]

Cooper et al.\ (1996)\cite{cooperCooperationReputationExperimental1996} & PD & Lab & 800 & 0 & 1000 & 350 & 0.30 \\[4pt]

Boone et al.\ (1999)\cite{booneImpactPersonalityBehavior1999a} & PD & Lab & 300 & $-$600 & 600 & $-$30 & 0.17 \\[4pt]

Swope et al.\ (2008)\cite{swopePersonalityPreferencesLaboratory2008} & PD & Lab & 10 & 2 & 20 & 5 & 0.46 \\[4pt]

Mengel (2018)\cite{mengelRiskTemptationMetastudy2018}$^{*}$ & PD & Online & 10 & 1 & 90 & 5 & 0.40 \\
 & PD & Lab    & 10 & 1 & 90 & 5 & 0.09 \\
 & PD & Online & 10 & 5 & 90 & 5 & 0.61 \\
 & PD & Online & 400 & 10 & 800 & 200 & 0.37 \\
 & PD & Online & 400 & 200 & 800 & 200 & 0.65 \\
 & PD & Lab    & 400 & 10 & 800 & 200 & 0.10 \\[4pt]

Gächter (2024)\cite{gachterRolePayoffParameters2024} & PD & Lab & 500 & 90 & 600 & 200 & 0.49 \\
 & PD & Lab & 500 & 20 & 600 & 200 & 0.45 \\
 & PD & Lab & 500 & 90 & 800 & 200 & 0.36 \\
 & PD & Lab & 500 & 20 & 800 & 200 & 0.38 \\
 & PD & Lab & 500 & 180 & 600 & 400 & 0.38 \\
 & PD & Lab & 500 & 40 & 600 & 400 & 0.33 \\
 & PD & Lab & 500 & 180 & 800 & 400 & 0.28 \\
 & PD & Lab & 500 & 40 & 800 & 400 & 0.28\\[4pt]

\midrule
\multicolumn{8}{l}{\textit{Prisoner's Dilemma --- repeated}} \\[2pt]

Rapoport \& Chammah (1965)\cite{rapoport_1965} & PD & Lab & 9 & $-$10 & 10 & $-$1 & -- \\
 & PD & Lab & 1 & $-$10 & 10 & $-$9 & -- \\
 & PD & Lab & 1 & $-$10 & 10 & $-$1 & -- \\
 & PD & Lab & 1 & $-$2 & 2 & $-$1 & -- \\
 & PD & Lab & 1 & $-$50 & 50 & $-$1 & -- \\
 & PD & Lab & 5 & $-$10 & 10 & $-$1 & -- \\
 & PD & Lab & 1 & $-$10 & 10 & $-$5 & -- \\[4pt]

Dal B\`{o} (2005)\cite{boCooperationShadowFuture2005} & PD & Lab & 65 & 10 & 100 & 35 & 0.13 \\
 & PD & Lab & 75 & 10 & 100 & 45 & 0.07 \\[4pt]

Bereby-Meyer \& Roth (2006)\cite{bereby-meyerSpeedLearningNoisy2006} & PD & Lab & 0.105 & 0.005 & 0.175 & 0.075 & -- \\[4pt]

Grimm \& Mengel (2009)\cite{grimmCooperationViscousPopulations2009} & PD & Lab & 800 & 100 & 850 & 150 & 0.62 \\
 & PD & Lab & 800 & 100 & 1100 & 400 & 0.10 \\[4pt]

Dal B\`{o} (2011) \cite{dalboEvolutionCooperationInfinitely2011} & PD & Lab & $R \in \{32, 40, 48\}$ & 12 & 50 & 25 & -- \\[4pt]

Dal B\`{o} \& Fr\'{e}chette (2019)\cite{dalboStrategyChoiceInfinitely2019} & PD & Lab & $R \in \{32, 48\}$ & 12 & 50 & 25 & -- \\[4pt]

\midrule
\multicolumn{8}{l}{\textit{Stag Hunt and Snowdrift games}} \\[2pt]

Capraro \& Halpern (2020)\cite{capraroPreferencesEfficiencyRather2020} & SH & Online & 40 & 5 & 25 & 25 & 0.63 \\[4pt]

K\"{u}mmerli et al.\ (2007)\cite{kummerliHumanCooperationSocial2007} & PD & Lab & 300 & $-$100 & 400 & 0 & 0.29 \\
 & SG & Lab & 200 & 100 & 300 & 0 & 0.48 \\[4pt]

Duffy \& Feltovich (2006)\cite{duffyWordsDeedsLies2006} & PD & Lab & 70 & 10 & 80 & 40 & 0.22 \\
 & SH & Lab & 70 & 10 & 55 & 55 & 0.61 \\
 & SG & Lab & 70 & 50 & 80 & 40 & 0.54 \\[4pt]

Kim \& Palfrey (2026)\cite{kimExperimentalStudyPrisoners2026} & PD & Lab & 60 & 11 & 98 & 35 & 0.16 \\
 & PD & Lab & 78 & 11 & 89 & 28 & 0.28 \\
 & SH & Lab & 77 & 13 & 75 & 65 & 0.30 \\
 & SH & Lab & 77 & 13 & 66 & 38 & 0.52 \\

\bottomrule
\end{tabular}
\begin{minipage}{\linewidth}
\vspace{8pt}
\footnotesize $^{*}$Mengel (2018) reports 13 payoff combinations across online (AMT 2013) and laboratory (Lab 2014) conditions; we show a representative selection. The same structural parameters (R\,=\,10, S\,=\,1, T\,=\,90, P\,=\,5) yield cooperation of 0.40 in the online setting and 0.09 in the laboratory, illustrating the sensitivity of outcomes to experimental context beyond payoff structure alone. The full set of combinations is reported in the original paper.
\end{minipage}
\end{table}